%% file: main.tex
\documentclass{article}

\usepackage{microtype}
\usepackage{graphicx}
\usepackage{subcaption}
\usepackage{booktabs} % for professional tables
\usepackage[table]{xcolor}
\usepackage{booktabs}
\usepackage{multirow}
\usepackage{graphicx}

\usepackage{hyperref}
\usepackage{xspace}

\newcommand{\method}{ST-VLA\xspace}
\newcommand{\methodset}{ST-Human\xspace}
\newcommand{\std}[1]{\textsubscript{\scalebox{0.9}{$\pm#1$}}}

\usepackage[preprint]{icml2026}

\usepackage{amsmath}
\usepackage{amssymb}
\usepackage{mathtools}
\usepackage{amsthm}

\usepackage[capitalize,noabbrev]{cleveref}

\theoremstyle{plain}

\theoremstyle{definition}

\theoremstyle{remark}

\usepackage[textsize=tiny]{todonotes}
\usepackage{multirow} 

\icmltitlerunning{ST-VLA}

\begin{document}

\twocolumn[
  \icmltitle{ST-VLA: Enabling 4D-Aware Spatiotemporal Understanding \\ for General Robot Manipulation}
  \icmlsetsymbol{equal}{*}

\begin{icmlauthorlist}
\icmlauthor{You Wu}{equal,nju-cs}
\icmlauthor{Zixuan Chen}{equal,nju-ai}
\icmlauthor{Cunxu Ou}{nju-ai}
\icmlauthor{Wenxuan Wang}{nju-ai}
\icmlauthor{Wenbo Huang}{nju-ee}
\icmlauthor{Lin Cao}{seu}
\icmlauthor{Yangtao Chen}{nju-cs}
\icmlauthor{Weichao Qiu}{huawei}
\icmlauthor{Xingyue Quan}{huawei}
\icmlauthor{Jieqi Shi}{nju-ai}
\icmlauthor{Jing Huo}{nju-cs}
\icmlauthor{Yang Gao}{nju-ai}
\end{icmlauthorlist}

\icmlaffiliation{nju-cs}{School of Computer Science, Nanjing University}
\icmlaffiliation{nju-ai}{School of Intelligence Science and Technology, Nanjing University}
\icmlaffiliation{nju-ee}{School of Electronic Science and Engineering, Nanjing University}
\icmlaffiliation{seu}{School of Instrument Science and Engineering, Southeast University}
\icmlaffiliation{huawei}{Noah's Ark Lab, Huawei}

\icmlcorrespondingauthor{You Wu}{you@smail.nju.edu.cn}

  % You may provide any keywords that you find helpful for describing your
  % paper; these are used to populate the "keywords" metadata in the PDF but
  % will not be shown in the document
  \icmlkeywords{Machine Learning, ICML}

  \vskip 0.3in
]
\renewcommand{\thefootnote}{\fnsymbol{footnote}}
\footnotetext[1]{Equal contribution.}
\renewcommand{\thefootnote}{\arabic{footnote}}

% \twocolumn[{%
% \renewcommand\twocolumn[1][]{#1}%
% \maketitle
% \vspace{-5mm}
% \begin{center}
%     \includegraphics[width=0.9\textwidth]{figs/teaserv2.pdf}
%     \vspace{-0.8mm}
%     \captionof{figure}{\textbf{ST-VLM} bridges the Semantic-Physical Gap via unified 3D-4D spatio-temporal representations. \textbf{(a)} Existing 2D-based VLMs face geometric ambiguity and temporal inconsistency due to the semantic-physical mismatch. \textbf{(Bottom)} Our ST-VLA utilizes unified 3D-4D representations with explicit trajectories and spatial masks, ensuring robust long-horizon manipulation.}
%     \label{fig:teaser}
%     % \vspace{-1mm}
% \end{center}
% }]

% this must go after the closing bracket ] following \twocolumn[ ...

% This command actually creates the footnote in the first column listing the
% affiliations and the copyright notice. The command takes one argument, which
% is text to display at the start of the footnote. The \icmlEqualContribution
% command is standard text for equal contribution. Remove it (just {}) if you
% do not need this facility.

% Use ONE of the following lines. DO NOT remove the command.
% If you have no special notice, KEEP empty braces:
\printAffiliationsAndNotice{}  % no special notice (required even if empty)
% Or, if applicable, use the standard equal contribution text:
% \printAffiliationsAndNotice{\icmlEqualContribution}

\begin{figure*}[t!]
    \centering
    \includegraphics[width=\textwidth]{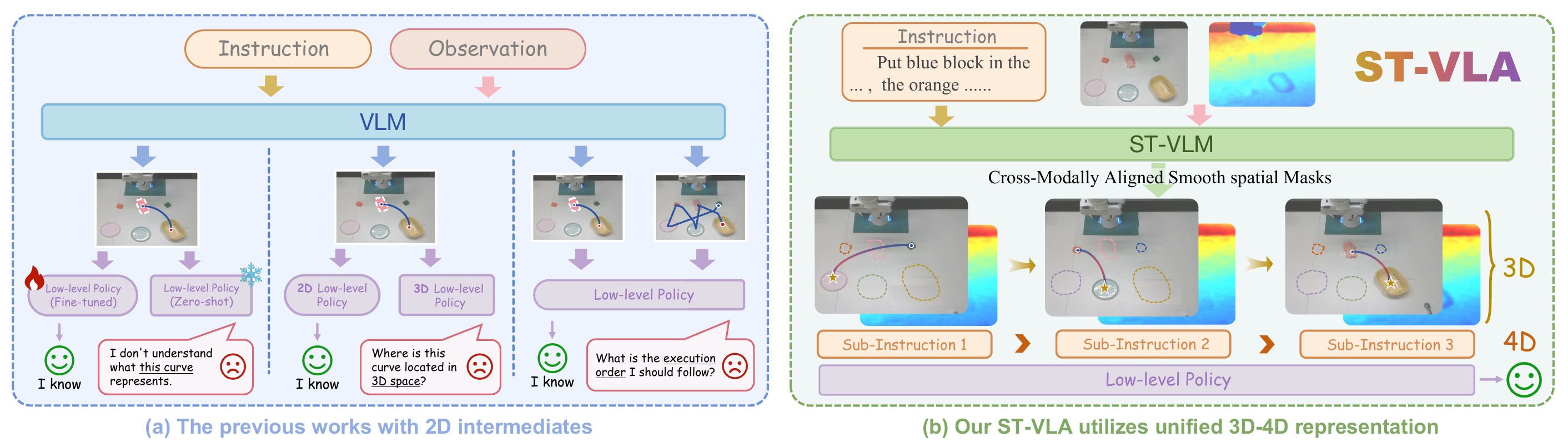}
    \caption{\textbf{ST-VLM} bridges the semantic-physical gap via unified 3D-4D spatio-temporal representations. \textbf{(Left)} Existing 2D-based VLMs face geometric ambiguity and temporal inconsistency due to the semantic-physical mismatch. \textbf{(Right)} Our \textbf{ST-VLA} utilizes unified 3D-4D representations with explicit trajectories and smooth spatial masks, ensuring robust long-horizon manipulation.}
    \label{fig:teaser}
    % \vspace{-5mm}
\end{figure*}

\input{sections/1_abs}
\input{sections/2_intro}

\input{sections/3_related_work}
\input{sections/4_method}
\input{sections/5_exp}
\input{sections/6_conclusion}

\bibliography{main}
\bibliographystyle{icml2026}

%%%%%%%%%%%%%%%%%%%%%%%%%%%%%%%%%%%%%%%%%%%%%%%%%%%%%%%%%%%%%%%%%%%%%%%%%%%%%%%
% APPENDIX
%%%%%%%%%%%%%%%%%%%%%%%%%%%%%%%%%%%%%%%%%%%%%%%%%%%%%%%%%%%%%%%%%%%%%%%%%%%%%%%
\newpage
\appendix
\onecolumn

\section*{Appendix Table of Contents}

\begin{itemize}
\item[\textbf{A}] \textbf{The Embodied Grasping Annotation Framework} \hfill \textbf{13}
\begin{itemize}
\item[A.1] System Architecture and Multi-Modal Viewport Synchronization \dotfill 13
\item[A.2] Action Semantic Templating and Output Schema \dotfill 14
\end{itemize}
\item[\textbf{B}] \textbf{Trajectory Extraction and Control Algorithms} \hfill \textbf{14}
\begin{itemize}
\item[B.1] 3D Trajectory Extraction and Processing Pipeline \dotfill 14
\item[B.2] Hierarchical Visual Masking for Long-Horizon Manipulation \dotfill 15
\end{itemize}
\item[\textbf{C}] \textbf{Single-Arm Task Suite Definitions} \hfill \textbf{16}
\begin{itemize}
\item[C.1] Data Collection Protocol and Environmental Diversity \dotfill 16
\item[C.2] Detailed Design of the 14 Single-Arm Tasks \dotfill 16
\end{itemize}
\item[\textbf{D}] \textbf{Dataset and Experimental Setup} \hfill \textbf{18}
\begin{itemize}
\item[D.1] PeractOneCam Benchmark Tasks \dotfill 18
\item[D.2] Dataset Statistics and Generalization Protocol \dotfill 18
\end{itemize}
\item[\textbf{E}] \textbf{Qualitative Visualizations} \hfill \textbf{18}
\begin{itemize}
\item[E.1] Pre-processing: SAM2 Segmentation \dotfill 18
\item[E.2] Evaluation: 2D Trajectory Generation (Ours vs GT) \dotfill 19
\end{itemize}
\item[\textbf{F}] \textbf{Training Prompts and Data Formats} \hfill \textbf{20}
\begin{itemize}
\item[F.1] Multimodal Prompt Templates \dotfill 20
\item[F.2] Unified Dataset JSON Representation \dotfill 20
\end{itemize}
\item[\textbf{G}] \textbf{Real-World Experimental Setup} \hfill \textbf{21} % 建议修改此处
\begin{itemize}
\item[G.1] Task Definitions \dotfill 21
\end{itemize}
\item[\textbf{H}] \textbf{Implementation Details} \hfill \textbf{22}
\begin{itemize}
\item[H.1] High-Level VLM Tuning (Qwen3-VL) \dotfill 22
\item[H.2] Low-Level Policy Training (3DDA and 3DFA) \dotfill 22
\end{itemize}
\item[\textbf{I}] \textbf{Detailed Experimental Results and Metric Definitions} \hfill \textbf{23}
\begin{itemize}
\item[I.1] 2D Tasks: Visual Grounding and Manipulation \dotfill 23
\item[I.2] 3D Tasks: Spatial Reasoning and Depth Estimation \dotfill 24
\item[I.3] 4D Tasks: Long-Horizon Planning \dotfill 24 
\end{itemize}
\end{itemize}

\newpage

\section{The Embodied Grasping Annotation Framework}
\label{app:annotation-framework}

\subsection{System Architecture and Multi-Modal Viewport Synchronization}
\begin{figure}[H]
\centering
% User preference: 0.80 width
\includegraphics[width=0.80\textwidth]{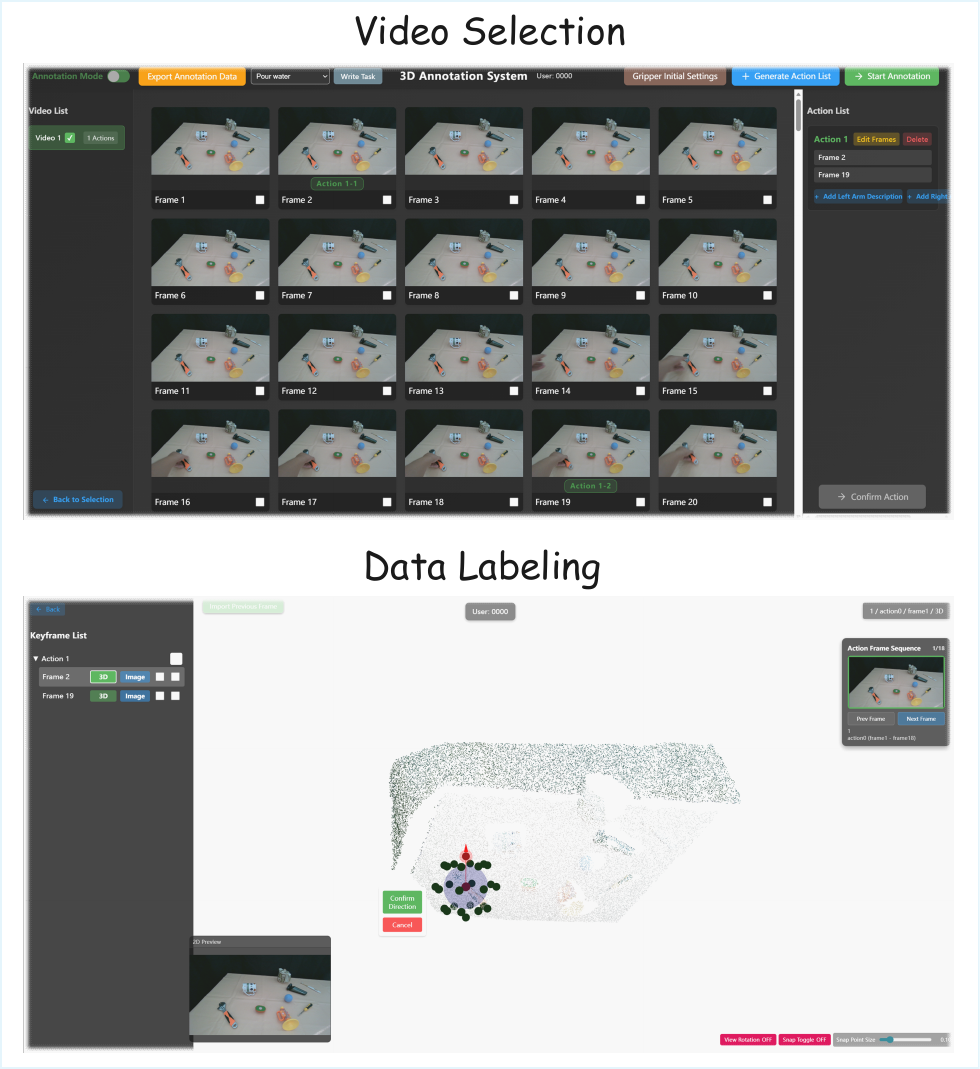}
\caption{Overview of the Annotation Software Workflow. \textbf{Top:} The \textit{Video Selection} interface for temporal segmentation, displaying frame sequences and action management lists. \textbf{Bottom:} The \textit{Data Labeling} interface featuring a central 3D point cloud editor (with manipulation gizmo), strictly synchronized with a 2D RGB preview window (bottom-left) to ensure spatial-visual alignment.}
\label{fig:annotation_example}
\end{figure}

Our specialized annotation suite is engineered using the React-Three-Fiber framework. As illustrated in Figure \ref{fig:annotation_example}, the workflow is divided into two operational stages to ensure precision:

\begin{enumerate}
    \item \textbf{Video Selection (Top Panel):} Annotators perform temporal segmentation using a frame-grid interface to define the start and end indices of atomic actions. The "Action List" sidebar allows for the management of complex, multi-step task sequences.
    
    \item \textbf{Data Labeling (Bottom Panel):} The core spatial annotation interface features a large \textbf{3D Point Cloud View} equipped with a 6-DoF gizmo for precise coordinate manipulation. Crucially, a \textbf{Synchronized 2D Preview} window (bottom-left) instantly renders the projection of 3D labels onto the RGB frame (\(P_{2D} = K[R|t]P_{3D}\)), providing closed-loop visual verification to minimize depth-perception errors.
\end{enumerate}

\begin{itemize}
    \item \textbf{Tier 00 (Annotation):} Operators define 6-DoF trajectory points, boolean gripper states ("0"=open, "1"=closed), and temporal boundaries.
    
    \item \textbf{Tier 01 (Audit):} A 100\% manual review process where senior experts verify every annotated segment for spatial accuracy (\(<5\) mm deviation) and temporal alignment.
    
    \item \textbf{Tier 10 (Acceptance):} Automated statistical sampling. A batch is only committed to the database if it achieves a \(>95\%\) pass rate in random checks.
    
    \item \textbf{Tier 11 (Admin):} Global configuration of user permissions and data export pipelines (JSON/Parquet).
\end{itemize}

\subsection{Action Semantic Templating and Output Schema}
We utilize a library of 18 structured action templates to convert manual labels into machine-readable language instructions. Algorithm \ref{alg:raw_json} illustrates the primary data structure (info.json) generated by our framework.

\begin{algorithm}[H]
\caption{Raw Annotation Data Output Schema}
\label{alg:raw_json}
\begin{algorithmic}[1]
\STATE \{
\STATE \ \ "task\_description": "Pouring Water",
\STATE \ \ "action\_descriptions": [ \{
\STATE \ \ \ \ "frame\_range": \{ "start\_frame": "frame1", "end\_frame": "frame8" \},
\STATE \ \ \ \ "left\_description": \{
\STATE \ \ \ \ \ \ "action\_description": "Pick up the coffee goblet",
\STATE \ \ \ \ \ \ "coordinate\_description": \{
\STATE \ \ \ \ \ \ \ \ "coordinate\_0": \{
\STATE \ \ \ \ \ \ \ \ \ \ "text": "Pick up the coffee goblet",
\STATE \ \ \ \ \ \ \ \ \ \ "image\_coordinates": [417, 170],
\STATE \ \ \ \ \ \ \ \ \ \ "cartesian\_coordinates": [-0.031, -0.115, 0.674, 0.153, 0.013, 0.633]
\STATE \ \ \ \ \ \ \ \ \ \} \}
\STATE \ \ \ \ \}
\STATE \ \ \} ],
\STATE \ \ "parameter\_data": \{ "camera\_intrinsics": "[[427.17, ...]]" \}
\STATE \}
\end{algorithmic}
\end{algorithm}

\newpage

\section{Trajectory Extraction and Control Algorithms}
\label{app:algorithms}

\subsection{3D Trajectory Extraction and Processing Pipeline}
The transformation of raw visual observations into smooth robotic actions is formalized in Algorithm \ref{alg:traj_extraction}.

\begin{algorithm}[H]
\caption{3D Trajectory Extraction and Processing Pipeline}
\label{alg:traj_extraction}
\begin{algorithmic}[1]
\STATE \textbf{Input:} Video \(V\), Depth \(D\), Initial Point \(p_{init}\)
\STATE \textbf{Step 1 (Tracking):} Initialize SAM2 on \(V\) with \(p_{init}\). Generate 2D track \(T_{2D}\). Apply linear interpolation for minor occlusions.
\STATE \textbf{Step 2 (Filtering):} Compute motion vector \(\vec{v}\). Discard points \(p_t\) where projection distance to \(\vec{v}\) exceeds threshold \(\epsilon\) (Outlier Rejection).
\STATE \textbf{Step 3 (Fusion):} Retrieve depth \(z \leftarrow D(u,v)\). Unproject to 3D space: \(P_{3D} = K^{-1} \cdot [u,v,1]^T \cdot z\).
\STATE \textbf{Step 4 (Smoothing):} Apply \texttt{WeightedPolyFit} (degree=2) to ensure kinematic feasibility.
\STATE \textbf{Step 5 (Sampling):} Downsample to \(K=8\) uniform keypoints for canonical representation \(\Omega\).
\end{algorithmic}
\end{algorithm}

While the spatial tube effectively filters the workspace, certain interactions require even finer granularity. To ensure precision at the interaction interface, we implement an endpoint proximity constraint. If no object is detected within a 10cm radius of the trajectory's terminal waypoint \(p_K\), the system automatically identifies the nearest object within this search domain as a relevant entity. This heuristic guarantees that the policy maintains focus on the target object or the destination of the manipulation.

\subsection{Hierarchical Visual Masking for Long-Horizon Manipulation}
To mitigate environmental distractions in multi-stage tasks, we employ Algorithm \ref{alg:visual_masking}.

\begin{algorithm}[H]
\caption{Hierarchical Visual Masking for Long-Horizon Focus}
\label{alg:visual_masking}
\begin{algorithmic}[1]
\STATE \textbf{Input:} Instruction List \(L\), Stage \(k\)
\WHILE{Task not finished}
\STATE \(\tau_{ref} \leftarrow \text{GetReferenceTrajectory}(\text{Stage } k)\)
\STATE \(M_{all} \leftarrow \text{SAM2Segment}(I_{rgb}, L[k])\)
\STATE \(M_{focus} \leftarrow \{m \in M_{all} \mid \text{Intersect}(m, \tau_{ref})\}\)
\STATE \(\hat{I}_{rgb} \leftarrow \text{Inpaint}(I_{rgb}, \text{Mask} = M_{all} \setminus M_{focus})\)
\STATE Execute Policy \(\pi(\hat{I}_{rgb})\) until stage completion criteria met.
\STATE \(k \leftarrow k+1\)
\ENDWHILE
\end{algorithmic}
\end{algorithm}

\newpage

\section{Single-Arm Task Suite Definitions}
\label{app:task-suite}

\subsection{Data Collection Protocol and Environmental Diversity}

Our dataset covers diverse manipulation primitives \textbf{including fluid pouring, cloth folding, and multi-object assembly}. Using a fixed RGB-D sensor, we capture kinematically feasible human demonstrations where the hand exits the view between modular grasp-and-release cycles to ensure clean temporal segmentation.

To promote robust open-world generalization, we systematically vary lighting, background textures, and object attributes, and introduce random distractors during collection. The primary axes of variation include:

\begin{enumerate}
    \item \textbf{Object Semantics:} Systematic variation of \textbf{color, material, and geometry} (e.g., stiffness of cables, opacity of bottles).
    
    \item \textbf{Pose Randomization:} Objects are instantiated with random \(SE(2)\) poses on the desktop surface.
    
    \item \textbf{Environmental Factors:} Integration of \(\geq 20\) desktop textures and randomized lighting conditions.
\end{enumerate}

\subsection{Detailed Design of the 14 Single-Arm Tasks}
This section details the operational design for the single-arm manipulation subset (approx. 800,000 frames).

\textbf{Task 1: Pouring Water (40k Frames)} \\
\textit{Objective:} Transfer liquid from a source container to a target container. \\
\textit{Core Variables:} Fluid interactions, container opacity, shape diversity (kettles, bottles, tea cups). \\
\textit{Procedure:} Grasp source object, translate to target vicinity, rotate until liquid is transferred, and return to upright.

\textbf{Task 2: Wiping Surfaces (40k Frames)} \\
\textit{Objective:} Clean a spill (liquid or powder) using a cleaning tool. \\
\textit{Core Variables:} Stain types (colored liquids, coffee powder), tool types (sponges, rags). \\
\textit{Procedure:} Retrieve tool, execute planar wiping motion (linear/circular) covering the stain area.

\textbf{Task 3: Stacking Cups (80k Frames)} \\
\textit{Objective:} Stack 5 paper cups into a stable pyramid structure. \\
\textit{Core Variables:} Cup deformation, texture variations (\(>10\) styles), initial scattering. \\
\textit{Procedure:} Precise pick-and-place with vertical alignment and gravity stability.

\textbf{Task 4: Building Blocks (80k Frames)} \\
\textit{Objective:} Assemble blocks into specified shapes (numbers, animals, trees). \\
\textit{Core Variables:} Block colors/shapes, target topology (requires \(\geq 5\) blocks). \\
\textit{Procedure:} Sequentially pick blocks from random piles and place them to form connected 2D/3D structures.

\textbf{Task 5: Pick and Place (Categorization) (110k Frames)} \\
\textit{Objective:} Sort objects into specific containers (boxes, drawers). \\
\textit{Core Variables:} Object categories (toys, food, stationery), container types (bins, shelves). \\
\textit{Procedure:} Identify object category, transport to matched container, and release safely.

\textbf{Task 6: Scene Cleaning (25k Frames)} \\
\textit{Objective:} Clear desktop by moving trash items to a bin. \\
\textit{Core Variables:} Trash types (crumpled paper, peel), bin location/type. \\
\textit{Procedure:} Grasp irregular trash objects and release them above the disposal zone.

\textbf{Task 7: Control Adjustment (40k Frames)} \\
\textit{Objective:} Operate mechanical controls such as valves, knobs, or buttons. \\
\textit{Core Variables:} Mechanism types (rotary valves, push buttons, switches). \\
\textit{Procedure:} Fine-grained manipulation: grasp handle and rotate (CW/CCW), or apply force to buttons.

\textbf{Task 8: Target Placement (150k Frames)} \\
\textit{Objective:} Place objects in specific relative poses (e.g., "to the left of X", "upside down"). \\
\textit{Core Variables:} Relative semantics (left/right/in/on), 6-DoF orientation requirements. \\
\textit{Procedure:} Pick object, re-orient gripper for target pose, and place according to spatial instruction.

\textbf{Task 9: Tower Stacking (50k Frames)} \\
\textit{Objective:} Stack \(\geq 4\) objects (books, boxes) vertically without collapse. \\
\textit{Core Variables:} Object friction, size ratios, center-of-mass estimation. \\
\textit{Procedure:} Iteratively stack items, ensuring alignment of centers to maintain tower balance.

\textbf{Task 10: Push and Pull (40k Frames)} \\
\textit{Objective:} Translate heavy or non-graspable objects across the surface. \\
\textit{Core Variables:} Surface friction, obstacle placement, object mass. \\
\textit{Procedure:} Apply non-prehensile force to slide objects to target zones, navigating around obstacles.

\textbf{Task 11: Point and Touch (100k Frames)} \\
\textit{Objective:} Semantically reference object parts (e.g., "touch the blue handle"). \\
\textit{Core Variables:} Object parts, color, semantic regions. \\
\textit{Procedure:} Move end-effector to close proximity of specific semantic regions without disrupting the object.

\textbf{Task 12: Material Cutting (50k Frames)} \\
\textit{Objective:} Use a tool to cut deformable material (Play-doh) at a specific ratio. \\
\textit{Core Variables:} Tools (knives), material consistency, cut ratios (1/2, 1/3). \\
\textit{Procedure:} Grasp tool, align blade with semantic cut point, and execute downward shearing force.

\textbf{Task 13: Flexible Object Interaction (45k Frames)} \\
\textit{Objective:} Manipulate deformable 1D objects (ropes, cables). \\
\textit{Core Variables:} Stiffness, length, entanglement state. \\
\textit{Procedure:} Straighten, fold, or insert flexible linear objects into containers.

\textbf{Task 14: Spatial Position Swap (20k Frames)} \\
\textit{Objective:} Swap locations of two objects. \\
\textit{Procedure:} Multi-step planning: (1) Move Object A to buffer; (2) Move Object B to A's origin; (3) Move Object A to B's origin.

\begin{figure}[H]
\centering
\includegraphics[width=0.45\textwidth]{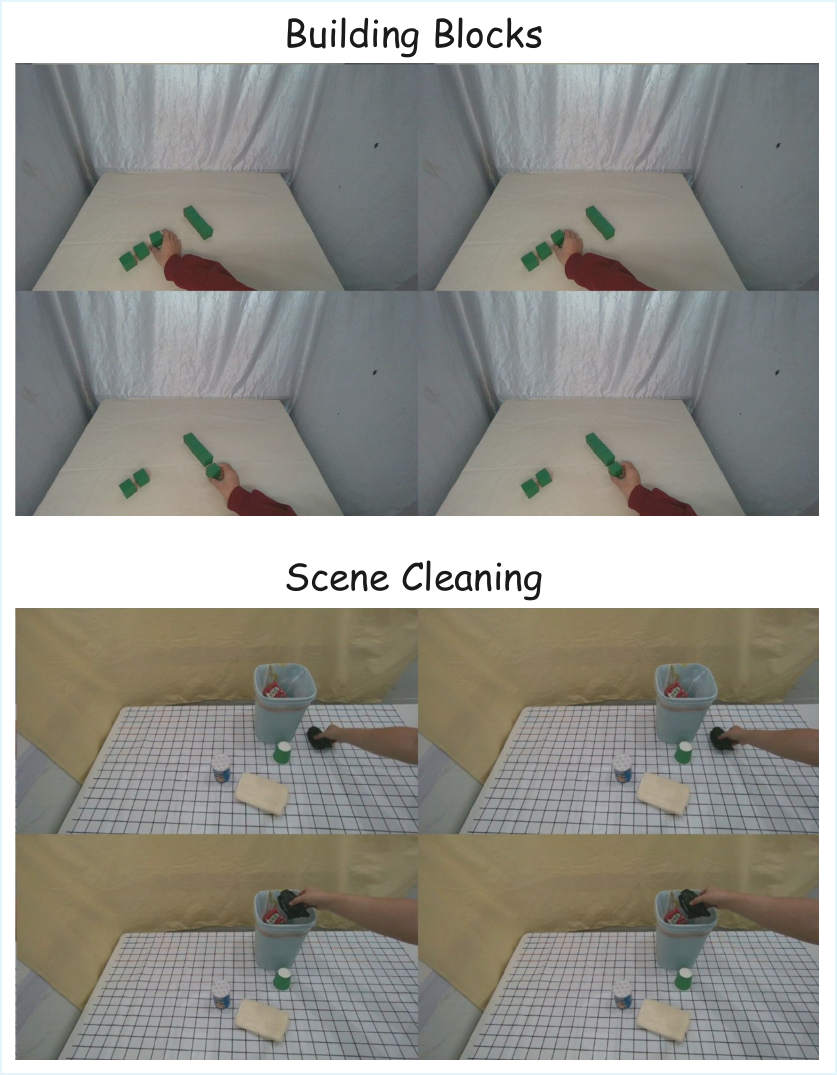}
\caption{Qualitative Examples of Dataset Tasks. \textbf{Top:} 'Building Blocks' task demonstrating multi-object spatial arrangement. \textbf{Bottom:} 'Scene Cleaning' task showing interaction with irregular trash objects. Both examples highlight the diversity in background textures and lighting conditions.}
\label{fig:data_examples}
\end{figure}

\newpage

\section{Dataset and Experimental Setup}
\label{app:dataset-setup}

\subsection{PeractOneCam Benchmark Tasks}
We evaluate our method on a subset of the PeractOneCam dataset, focusing on four challenging long-horizon manipulation tasks:

 \textbf{Close a jar:} There are two colored jars. The jar colors are sampled from a set of 20 colors. The agent needs to pick up the lid and screw it onto the jar with the specified color. The task involves six keyposes.
 \textbf{Screw a light bulb:} There are 2 light bulbs, 2 holders, and 1 lamp stand. The holder colors are sampled from a set of 20 colors. The agent needs to pick the correct light bulb and screw it into the matching colored holder.
 \textbf{Put groceries in the cupboard:} There are 9 YCB objects and a cupboard. The agent needs to grab the specified object and place it in the cupboard. The task involves 5 keyposes.
 \textbf{Push a button:} There are 3 buttons, whose colors are sampled from a set of 20 colors. The agent needs to push the colored buttons in the specified sequence. The task involves 3.8 keyposes on average.

\subsection{Dataset Statistics and Generalization Protocol}
The dataset consists of expert trajectories collected in RLBench, with \textbf{100 episodes per task} for training. To assess generalization capabilities, we adopt a \textbf{Seen/Unseen} split protocol (Table~\ref{tab:dataset_stats}) based on visual and semantic variations.

\textbf{Task Configurations and Generalization Rules.} 
For \textit{Close Jar} and \textit{Light Bulb In}, 5 of 20 color variations are held out to test zero-shot visual generalization. In \textit{Put Groceries}, 2 of 9 object instances (e.g., Crackers, Soup) are excluded to evaluate semantic category transfer. For \textit{Push Buttons}, we re-organize the original 50 variations—which involve sequential presses (1–3 times)—by decomposing them into single-press episodes. Specifically, trajectories are segmented every two keyposes, resulting in 78 variations across diverse button colors. This ensures each episode consists of exactly two keyposes, maximizing the model's ability for precise language-to-color grounding. 

\begin{table}[h]
\centering
\caption{Dataset Statistics and Variation Splits. \textit{Seen} indicates variations used for training (100 episodes/task); \textit{Unseen} indicates variations held out for generalization testing.}
\label{tab:dataset_stats}
\small
\setlength{\tabcolsep}{4pt}
\begin{tabular}{lcccc}
\toprule
\textbf{Task Name} & \textbf{Total Vars} & \textbf{Seen} & \textbf{Unseen} & \textbf{Excluded Examples (Unseen)} \\
\midrule
Close Jar & 20 & 15 & 5 & Specific Colors (e.g., Azure, Violet) \\
Light Bulb In & 20 & 15 & 5 & Specific Colors (e.g., Black, White) \\
Put Groceries & 9 & 7 & 2 & Novel Objects (e.g., Crackers, Soup) \\
Push Buttons & 78 & 78 & 0 & None (Fine-grained Color Training) \\
\bottomrule
\end{tabular}
\end{table}

\newpage

\section{Qualitative Visualizations}
\label{app:visualizations}

This section provides qualitative examples of our pipeline's intermediate processing (segmentation) and final model output (trajectory generation).

\subsection{Pre-processing: SAM2 Segmentation}
\label{app:visualizations_sam2}
We visualize the intermediate segmentation masks produced by SAM2 on three tasks. The robust mask generation is a critical precursor for accurate 3D lifting.

\begin{figure}[H]
\centering
% User preference: 0.70 width
\includegraphics[width=0.70\textwidth]{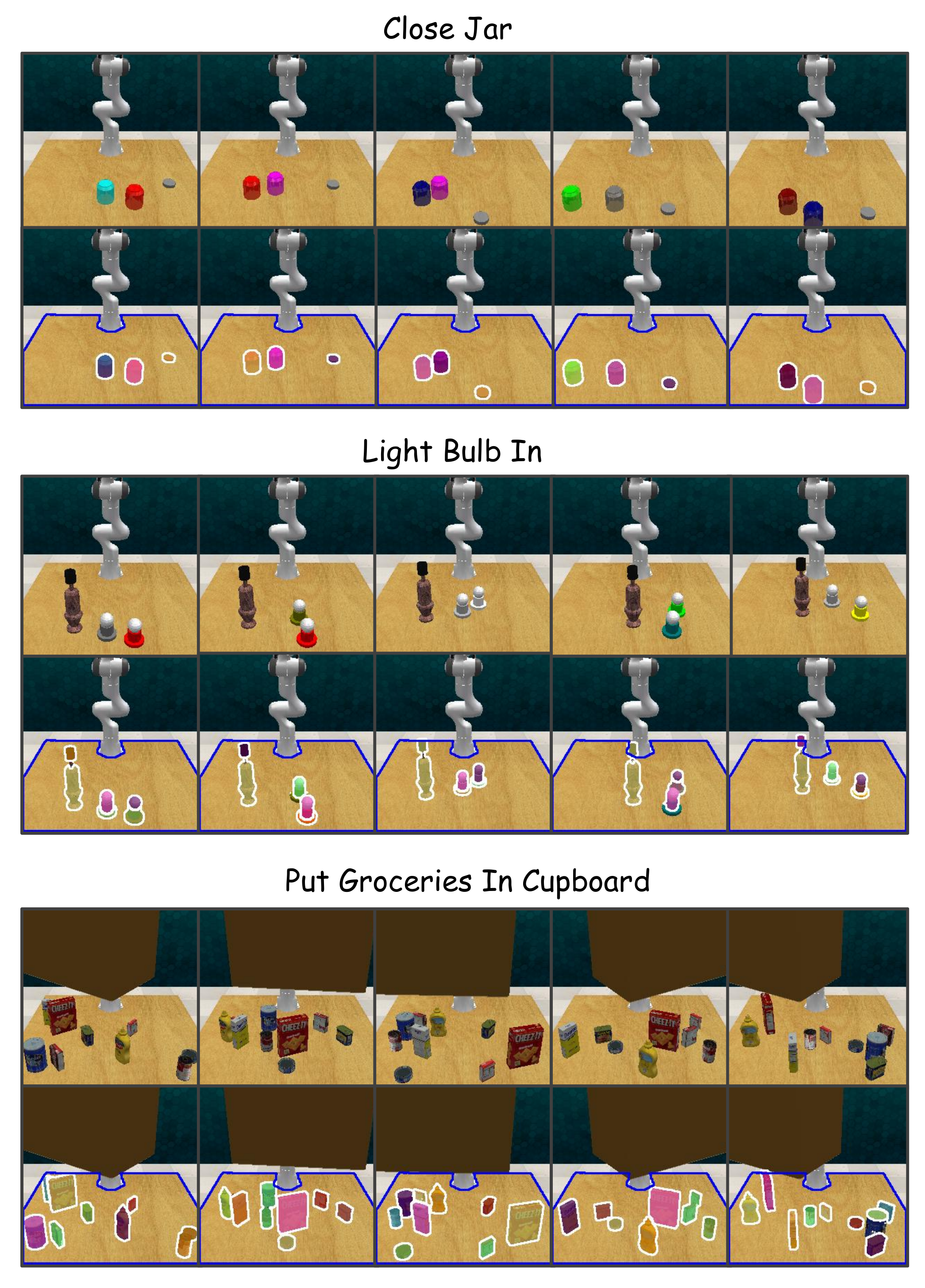}
\caption{Qualitative Visualization of SAM2 Segmentation. \textbf{Top:} 'Close Jar' task, showing precise masks for colored jars and loose lids. \textbf{Middle:} 'Light Bulb In' task, isolating distinct components. \textbf{Bottom:} 'Put Groceries In Cupboard' task, demonstrating robust segmentation in clutter.}
\label{fig:sam2_example}
\end{figure}

\subsection{Evaluation: 2D Trajectory Generation (Ours vs GT)}
\label{app:visualizations_traj}

\begin{figure}[H]
\centering
% New Figure from user, setting width to 0.95\textwidth as discussed
\includegraphics[width=0.75\textwidth]{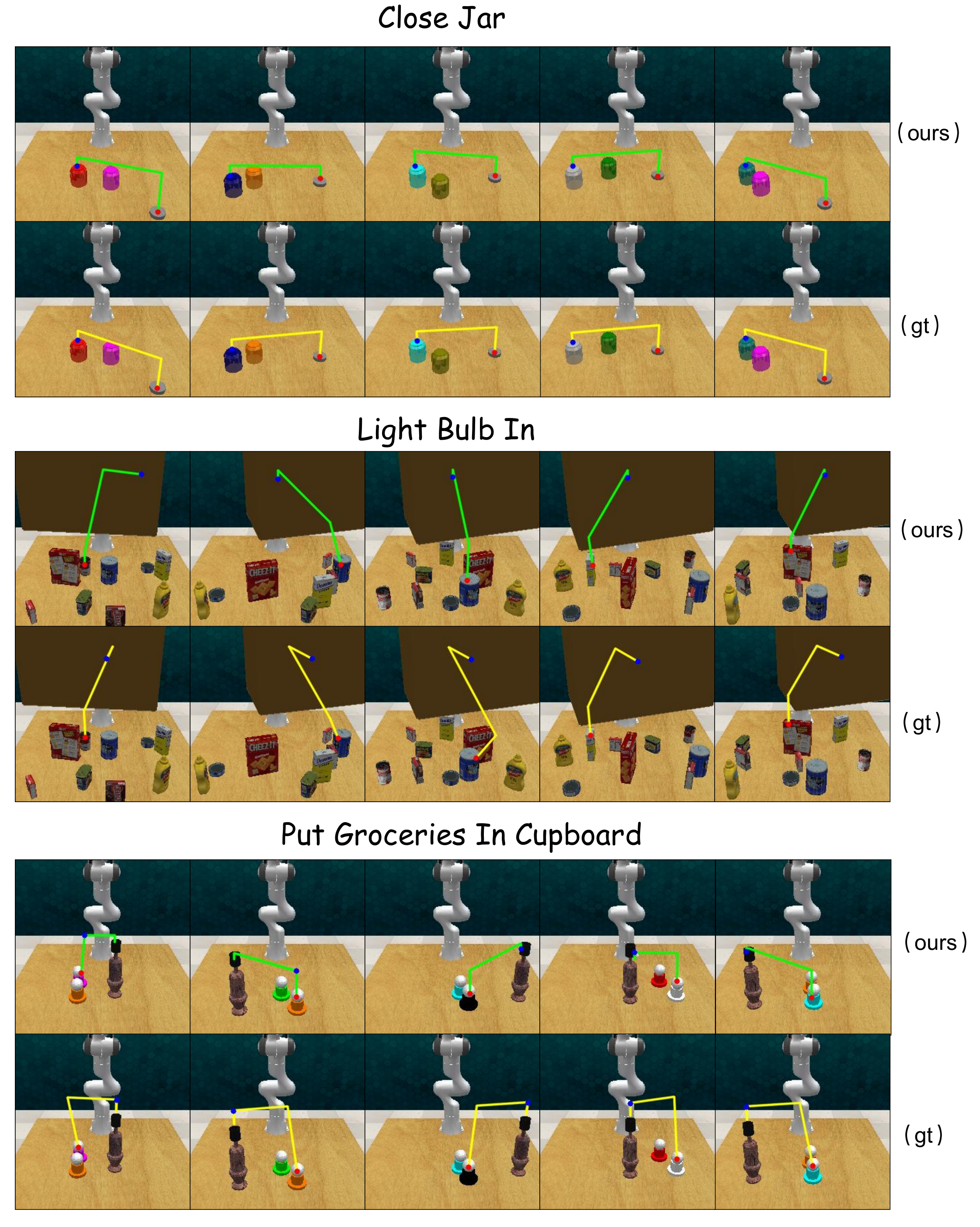}
\caption{Qualitative Comparison of Trajectory Generation. We visualize the 2D trajectories generated by our model (top row) versus ground truth human demonstrations (bottom row) across three tasks: Close Jar (Left), Light Bulb In (Middle), and Put Groceries In Cupboard (Right). The model successfully captures the critical keypoints and temporal geometry of the manipulation paths.}
\label{fig:traj_comparison}
\end{figure}

Figure \ref{fig:traj_comparison} compares the 2D trajectories generated by our model against the ground truth (GT) human demonstrations.

\section{Training Prompts and Data Formats}
\label{app:prompts-format}

\subsection{Multimodal Prompt Templates}
We categorize training prompts into spatial, temporal, and reasoning tasks:
\begin{itemize}
    \item \textbf{Trajectory:} \textit{"Instruction: '\textless task\textgreater'. Provide a sequence of exactly 8 points denoting the trajectory."}
    
    \item \textbf{2D Grounding:} \textit{"Pinpoint a 2D point within \textless object\textgreater normalized to [0-1000]."}
\end{itemize}

\subsection{Unified Dataset JSON Representation}
Table \ref{tab:json_format} shows the data structure used for fine-tuning.

\begin{table}[h!]
\centering
\caption{Unified Dataset JSON Format}
\label{tab:json_format}
\begin{small}
\begin{verbatim}
{
"messages": [
{ "role": "user", "content": "<image>Task: 'Wipe table'." },
{ "role": "assistant", "content": "<bbox>...<bbox>" }
],
"images": ["data/0001/rgb/frame_001.jpg"],
"objects": {
"traj_2d": [[500, 499], [516, 488], [534, 479], [552, 472],
[571, 466], [590, 462], [611, 460], [631, 459]]
}
}
\end{verbatim}
\end{small}
\end{table}

\section{Experimental Setup for Real World}
\label{app:real-world-setup}

\textbf{Robot Setup}. We conduct real-world experiments using a Franka Emika Panda robot arm, a 7-DoF manipulator equipped with a standard parallel-jaw gripper. To perceive the environment, we utilize a third-person front-view camera positioned on a tripod facing the workspace. This camera provides a global perspective of the tabletop and the manipulation targets. The robot is controlled via a workstation running Linux, communicating with the robot controller in real-time.

\textbf{Task Definitions}

\noindent \textbf{(a) pick\_place}

\noindent \textbf{Task Description:} The robot must grasp a target object (colored block or other item) and place it securely into a plate. \\
\textbf{Success Metric:} The task is successful if the object rests stably inside the boundaries of the plate after release. \\
\textbf{Objects:} Target object (variable color/type), one plate. \\
\textbf{Language Instructions:} \textit{Put the \textless color \textgreater block in the plate.} (or \textit{Put the \textless object \textgreater in the plate.}) \\
\textbf{Variation Number:} 4

\vspace{1em}

\noindent \textbf{(b) stack\_cube}

\noindent \textbf{Task Description:} The robot must pick up a specific colored block and stack it on top of another block. \\
\textbf{Success Metric:} The task is successful if the top block is stably aligned on the bottom block without toppling for 5 seconds. \\
\textbf{Objects:} Two colored blocks. \\
\textbf{Language Instructions:} \textit{Stack the \textless color1 \textgreater block on the \textless color2 \textgreater block.} \\
\textbf{Variation Number:} 6

\begin{table*}[h]
\centering
\caption{Properties of atomic tasks in real-world experiments.}
\label{tab:atomic_task_properties}
\resizebox{\textwidth}{!}{%
\begin{tabular}{@{}l l c c@{}} % 这里的列格式由 l l c c c 改为了 l l c c
    \toprule
    \multicolumn{1}{c}{Task name} & \multicolumn{1}{c}{Language Template} &  Variations & Variation Type \\
    \midrule
    pick\_place & ``Put the $<$color$>$ block in the plate.'' & 4 & Color, Object Type \\
     & \textit{(or ``Put the $<$object$>$ in the plate.'')} & & \\
    
    stack\_cube & ``Stack the $<$color1$>$ block on the $<$color2$>$ block.'' & 6 & Color Pair \\
    \bottomrule
\end{tabular}%
}
\end{table*}

\newpage

\section{Implementation Details}
\label{app:implementation}

\subsection{High-Level VLM Tuning (Qwen3-VL)}
We utilize the Qwen/Qwen3-VL-4B-Instruct model as our high-level planner. Table \ref{tab:qwen3_vlsft_params} details the hyperparameters used for the Supervised Fine-Tuning (SFT) process, where we employ LoRA targeting all linear modules to ensure efficient adaptation while preserving the pre-trained capabilities.

\subsection{Low-Level Policy Training (3DDA and 3DFA)}
We provide the detailed hyperparameters used for training our 3DDA and 3DFA models in Table \ref{tab:hyperparams}. The 3DDA model utilizes a DDPM-based diffusion process with 100 timesteps, employing a CLIP backbone for robust visual feature extraction. In contrast, the 3DFA model adopts a Rectified Flow formulation with only 5 denoising timesteps to enable rapid inference while maintaining generation quality. Both models are trained for 300,000 iterations with a batch size of 64.

\begin{table*}[h]
\centering
\caption{Hyperparameters for Qwen3-VL SFT Training}
\label{tab:qwen3_vlsft_params}
\begin{tabular}{ll}
\toprule
\textbf{Parameter} & \textbf{Value} \\
\midrule
\multicolumn{2}{l}{\textit{Training Optimization}} \\
Model & Qwen/Qwen3-VL-4B-Instruct \\
Training Type & LoRA \\
Batch Size (Per Device) & 6 \\
Gradient Accumulation Steps & 2 \\
Total Batch Size & \(6 \times 2 \times 8 = 96\) \\
Learning Rate & \(1 \times 10^{-4}\) \\
Weight Decay & Default (not specified) \\
Warmup Ratio & 0.05 \\
Num Epochs & 1 \\
Mixed Precision & bfloat16 \\
DeepSpeed Strategy & ZeRO-2 \\
\midrule
\multicolumn{2}{l}{\textit{LoRA Configuration}} \\
LoRA Rank & 64 \\
LoRA Alpha & 128 \\
Target Modules & all-linear \\
Freeze ViT & True \\
Freeze Aligner & True \\
\midrule
\multicolumn{2}{l}{\textit{Multimodal \& Sequence}} \\
Max Length & 4096 \\
Max Image Tokens & 1024 \\
Max Video Tokens & 128 \\
Max Video Frames (FPS) & 16 \\
Packing & True \\
Padding Free & True \\
Attention Implementation & Flash Attention \\
\midrule
\multicolumn{2}{l}{\textit{Dataset \& System}} \\
Validation Split Ratio & 0.01 \\
Num Workers (Dataloader) & 4 \\
Dataset Processes & 8 \\
Checkpoint Limit & 10 \\
Evaluation/Save Steps & 50 \\
\bottomrule
\end{tabular}
\end{table*}

\begin{table*}[h]
\centering
\caption{Hyperparameters for Peract Training and Evaluation}
\label{tab:hyperparams}
\begin{tabular}{ll}
\toprule
\textbf{Parameter} & \textbf{Value} \\
\midrule
\multicolumn{2}{l}{\textit{Training Optimization}} \\
Batch Size & 64 \\
Learning Rate & \(1 \times 10^{-4}\) \\
Backbone Learning Rate & \(1 \times 10^{-6}\) \\
Weight Decay & \(1 \times 10^{-10}\) \\
LR Scheduler & Constant \\
Training Iterations & 300,000 \\
\midrule
\multicolumn{2}{l}{\textit{Model Architecture}} \\
Model Type & Denoise3D \\
Backbone & CLIP \\
Embedding Dimension & 120 \\
Attention Heads & 8 \\
Vis-Instr Attention Layers & 2 \\
Shared Attention Layers & 4 \\
History Length & 3 frames \\
FPS Subsampling Factor & 4 \\
Rotation Format & Quaternion (xyzw) \\
\midrule
\multicolumn{2}{l}{\textit{Diffusion / Denoising}} \\
Denoise Model for 3DDA & DDPM \\
Denoise Timesteps for 3DDA & 100 \\
Denoise Model for 3DFA & RectifiedFlow \\
Denoise Timesteps for 3DFA & 5 \\
Prediction Length & 1 (Keypose Only) \\
\midrule
\multicolumn{2}{l}{\textit{Evaluation Environment}} \\
Image Size & \(256 \times 256\) \\
Max Steps & 25 \\
Max Tries (Motion Planner) & 1 \\
Dataset & PeractOneCam (front) \\
Seed & 0 - 2 \\
\bottomrule
\end{tabular}
\end{table*}

\newpage
\section{Detailed Experimental Results and Metric Definitions}
\label{sec:appendix_results}

In this appendix, we provide a comprehensive breakdown of the experimental results across 2D, 3D, and 4D benchmarks. We evaluate models on both standard public datasets and our collected \textbf{ST-Human} suite. The detailed metric definitions are as follows:

\subsection{2D Tasks: Visual Grounding and Manipulation}
The 2D tasks evaluate the model's capability in precise visual pointing and fine-grained trajectory generation. The quantitative results are shown in Table~\ref{tab:2d_results}.
\begin{itemize}
    \item \textbf{Hit Rate ($\uparrow$):} For standard pointing tasks, we calculate the hit rate based on ground truth annotations. 
    \begin{itemize}
        \item \textbf{Point-in-Box:} Used for \textit{RoboRefit} and \textit{VABench-Point}~\cite{yuan2025seeing}. A prediction is correct if it falls within the ground-truth bounding box.
        \item \textbf{Point-in-Mask:} Used for \textit{Where2Place} and \textit{Part-Afford}~\cite{myers2015affordance} A prediction is correct if it falls within the target segmentation mask.
    \end{itemize}
    \item \textbf{ST-Human-Pointing Metrics:} For our self-collected pointing dataset, we evaluate both the \textbf{Mean Euclidean Distance (MED $\downarrow$)} and the \textbf{Success Rate (SR $\uparrow$)}. Success is defined as the proportion of samples where the prediction error is below a pre-defined pixel threshold. (Table~\ref{tab:2d_results} reports SR).
    \item \textbf{Trajectory Metrics ($\downarrow$):} For trajectory generation tasks (\textit{VABench-Vision}, \textit{RLBench}, \textit{ST-Human-Trajectory}), we report the \textbf{Root Mean Square Error (RMSE)} and \textbf{Mean Absolute Error (MAE)}. Lower values indicate trajectories closer to human demonstrations.
\end{itemize}

% ==========================================================================================
% Table 1: 2D Tasks
% ==========================================================================================
\begin{table*}[h!]
\centering
\small
\caption{\textbf{Evaluation on 2D Tasks.} Metric types include: \textbf{Box-Hit} (RoboRefit, VAB-P), \textbf{Mask-Hit} (W2Pl, Afford), and \textbf{Success Rate} (ST-Human-Point). For trajectory tasks, we report RMSE ($\downarrow$).}
\label{tab:2d_results}
\setlength{\tabcolsep}{5pt}
\renewcommand{\arraystretch}{1.2}
\resizebox{\textwidth}{!}{%
\begin{tabular}{l ccccc ccc}
\toprule
\textbf{Method} & \textbf{RoboRefit} & \textbf{Where2Place} & \textbf{VABench-P} & \textbf{Part-Afford} & \textbf{ST-Human-Pointing} & \textbf{VABench-V} $\downarrow$ & \textbf{RLBench} $\downarrow$ & \textbf{ST-Human-Traj} $\downarrow$ \\
\midrule
% --- Closed-source ---
\rowcolor{groupgray} \multicolumn{9}{l}{\textit{Closed-source models}} \\
GPT-5.2         & 16.65\% & 43.00\% & 60.33\% & 23.50\% & 20.50\% & 174.00 & 66.34 & 130.08 \\
Gemini-3-flash  & 57.75\% & \textbf{87.00\%} & 61.00\% & \textbf{86.00\%} & 75.50\% & 122.00 & 72.82 & 42.36 \\
% --- Open-source ---
\rowcolor{groupgray} \multicolumn{9}{l}{\textit{Open-source models}} \\
Embodied-R1-3B     & 85.05\% & 66.83\% & \textbf{73.50\%} & 44.74\% & 61.00\% & 72.10 & 65.89 & 78.85 \\
PEEK-3B         & 10.05\% & 14.00\% & 17.33\% & 8.95\%  & 3.00\%  & 188.89 & 81.59 & 193.04 \\
Robobrain2.0-3B    & 50.61\% & 61.00\% & 26.88\% & 28.35\% & 65.50\% & 153.86 & 72.98 & 71.76 \\
Qwen3v1-4B      & 83.63\% & 65.00\% & 23.17\% & 28.40\% & 27.50\% & 191.90 & 95.80 & 348.50 \\
% --- Ours ---
\midrule
\rowcolor{bestblue} 
\textbf{ST-VLA (Ours)} & \textbf{88.15\%} & 73.00\% & 59.67\% & 49.40\% & \textbf{96.50\%} & \textbf{70.65} & \textbf{41.27} & \textbf{16.91} \\
\bottomrule
\end{tabular}%
}
\end{table*}

\subsection{3D Tasks: Spatial Reasoning and Depth Estimation}
Table~\ref{tab:3d_results} presents the results for 3D understanding.
\begin{itemize}
    \item \textbf{QA Accuracy ($\uparrow$):} For spatial relationship tasks (\textit{CVBench}, \textit{CRPE}, \textit{SAT}, \textit{Blink}, \textit{ST-Human-Spatial}), the model is evaluated in a multiple-choice format. We report the standard accuracy. \textit{Notably, ST-VLA exhibits a performance gap on CRPE compared to CVBench and SAT. We attribute this to CRPE's emphasis on \textbf{fine-grained geometric perception} (e.g., precise depth ordering and occlusion), whereas datasets like CVBench focus more on \textbf{high-level semantic spatial relationships}, which align better with the action-oriented nature of our model.}
    \item \textbf{Depth Metrics:} For the \textit{ST-Human-Depth} task:
    \begin{itemize}
        \item \textbf{Ratio Accuracy ($\uparrow$):} The proportion of samples where the predicted depth error is within $20\%$ of the ground truth value.
        \item \textbf{MAD ($\downarrow$):} The Mean Absolute Deviation in centimeters.
    \end{itemize}
\end{itemize}

% ==========================================================================================
% Table 2: 3D Tasks
% ==========================================================================================
\begin{table*}[h!]
\centering
\small
\caption{\textbf{Evaluation on 3D Tasks.} Spatial tasks use multiple-choice accuracy. Depth estimation reports Ratio Accuracy (threshold 20\%) and MAD.}
\label{tab:3d_results}
\setlength{\tabcolsep}{6pt}
\renewcommand{\arraystretch}{1.2}
\resizebox{\textwidth}{!}{%
\begin{tabular}{l cccccccc}
\toprule
\multirow{2}{*}{\textbf{Method}} & \multirow{2}{*}{\textbf{CVBench}} & \multirow{2}{*}{\textbf{CRPE}} & \multicolumn{2}{c}{\textbf{SAT}} & \multirow{2}{*}{\textbf{Blink}} & \multirow{2}{*}{\textbf{ST-Human-Spatial}} & \multicolumn{2}{c}{\textbf{ST-Human-Depth}} \\
\cmidrule(lr){4-5} \cmidrule(lr){8-9}
& & & Val & Test & & & Ratio Acc $\uparrow$ & MAD(cm) $\downarrow$ \\
\midrule
% --- Closed-source ---
\rowcolor{groupgray} \multicolumn{9}{l}{\textit{Closed-source models}} \\
GPT-5.2         & 79.62\% & 78.89\% & 66.56\% & 66.00\% & 49.08\% & 68.00\% & 4.00\% & 12.22 \\
Gemini-3-flash  & 83.94\% & \textbf{80.21\%} & 62.63\% & 64.67\% & 56.23\% & 84.00\% & 6.00\% & 13.82 \\
% --- Open-source ---
\rowcolor{groupgray} \multicolumn{9}{l}{\textit{Open-source models}} \\
Embodied-R1-3B     & 81.69\% & 74.78\% & 74.66\% & 70.00\% & \textbf{62.26\%} & 56.00\% & 2.70\% & 54.00 \\
PEEK-3B         & 55.49\% & 64.55\% & 50.76\% & 52.67\% & 40.35\% & 50.00\% & 0.00\% & 10.00 \\
Robobrain2.0-3B    & 81.22\% & 73.03\% & 59.64\% & 69.33\% & 51.24\% & 48.00\% & 8.89\% & 10.31 \\
Qwen3v1-4B      & 79.21\% & 77.89\% & 65.75\% & 68.67\% & 62.13\% & 62.00\% & 9.33\% & 14.07 \\
% --- Ours ---
\midrule
\rowcolor{bestblue} 
\textbf{ST-VLA (Ours)} & \textbf{84.52\%} & 73.67\% & \textbf{93.48\%} & \textbf{75.33\%} & 50.00\% & \textbf{98.00\%} & \textbf{46.67\%} & \textbf{1.82} \\
\bottomrule
\end{tabular}%
}
\end{table*}

\subsection{4D Tasks: Long-horizon Planning}
Table~\ref{tab:4d_results} details the evaluation on the \textit{ST-Human-Planning} dataset.
\begin{itemize}
    \item \textbf{Instruction Prediction ($\uparrow$):} For predicting the \textit{Next} and \textit{Previous} instructions, we compute the \textbf{BERTScore} to measure semantic similarity with ground-truth texts.
    \item \textbf{Task Progress Estimation:}
    \begin{itemize}
        \item \textbf{Step Acc ($\uparrow$):} The proportion of samples where the predicted step index exactly matches the ground truth.
        \item \textbf{Step MAE ($\downarrow$):} The average absolute deviation of the predicted step index.
        \item \textbf{Status Acc ($\uparrow$):} A binary classification accuracy measuring if the model correctly identifies whether the task has finished.
    \end{itemize}
    \item \textbf{Next-Step Trajectory ($\downarrow$):} We evaluate the planned future trajectory using \textbf{RMSE} and \textbf{MAE}.
\end{itemize}

% ==========================================================================================
% Table 3: 4D Tasks
% ==========================================================================================
\begin{table*}[h!]
\centering
\small
\caption{\textbf{Evaluation on 4D Tasks (ST-Human-Planning).} Evaluating instruction generation (BERTScore), progress estimation (Step/Status metrics), and future planning error.}
\label{tab:4d_results}
\setlength{\tabcolsep}{8pt}
\renewcommand{\arraystretch}{1.2}
\resizebox{\textwidth}{!}{%
\begin{tabular}{l cc ccc cc}
\toprule
\multirow{2}{*}{\textbf{Method}} & \textbf{Next Act} & \textbf{Prev Act} & \multicolumn{3}{c}{\textbf{Progress Estimation}} & \multicolumn{2}{c}{\textbf{Plan Error} $\downarrow$} \\
\cmidrule(lr){2-2} \cmidrule(lr){3-3} \cmidrule(lr){4-6} \cmidrule(lr){7-8}
& BERT $\uparrow$ & BERT $\uparrow$ & Step Acc $\uparrow$ & Step MAE $\downarrow$ & Status Acc $\uparrow$ & RMSE & MAE \\
\midrule
% --- Closed-source ---
\rowcolor{groupgray} \multicolumn{8}{l}{\textit{Closed-source models}} \\
GPT-5.2         & 0.89 & 0.88 & 12.00\% & 4.62 & 88.0\% & 199.35 & 186.77 \\
Gemini-3-flash  & 0.83 & 0.82 & 16.00\% & 4.10 & 84.0\% & 283.47 & 245.50 \\
% --- Open-source ---
\rowcolor{groupgray} \multicolumn{8}{l}{\textit{Open-source models}} \\
Embodied-R1-3B     & 0.79 & 0.78 & 14.00\% & 4.48 & 78.0\% & 129.94 & 76.86 \\
PEEK-3B         & 0.77 & 0.80 & 14.00\% & 4.90 & 88.0\% & 178.29 & 109.78 \\
Robobrain2.0-3B    & 0.88 & 0.88 & 18.00\% & 4.50 & 88.0\% & 256.94 & 253.11 \\
Qwen3v1-4B      & 0.77 & 0.70 & 14.00\% & 4.78 & 88.0\% & 448.75 & 394.96 \\
% --- Ours ---
\midrule
\rowcolor{bestblue} 
\textbf{ST-VLA (Ours)} & \textbf{0.97} & \textbf{0.98} & \textbf{60.00\%} & \textbf{0.70} & \textbf{92.0\%} & \textbf{43.66} & \textbf{24.13} \\
\bottomrule
\end{tabular}%
}
\end{table*}
% ----------------------------------------

\end{document}

%% file: sections/1_abs.tex
\begin{abstract}
Robotic manipulation in open-world environments requires reasoning across semantics, geometry, and long-horizon action dynamics. Existing hierarchical Vision-Language-Action (VLA) frameworks typically use 2D representations to connect high-level reasoning with low-level control, but lack depth awareness and temporal consistency, limiting robustness in complex 3D scenes.
We propose \textbf{\method}, a hierarchical VLA framework using a unified 3D-4D representation to bridge perception and action. \method converts 2D guidance into 3D trajectories and generates smooth spatial masks that capture 4D spatio-temporal context, providing a stable interface between semantic reasoning and continuous control. 
To enable effective learning of such representations, we introduce \textbf{ST-Human}, a large-scale human manipulation dataset with 14 tasks and 300k episodes, annotated with 2D, 3D, and 4D supervision via a semi-automated pipeline. Using ST-Human, we train \textbf{ST-VLM}, a spatio-temporal vision-language model that generates spatially grounded and temporally coherent 3D representations to guide policy execution. The smooth spatial masks focus on task-relevant geometry and stabilize latent representations, enabling online replanning and long-horizon reasoning.
Experiments on RLBench and real-world manipulation tasks show that \method significantly outperforms state-of-the-art baselines, improving zero-shot success rates by 44.6\% and 30.3\%. These results demonstrate that offloading spatio-temporal reasoning to VLMs with unified 3D-4D representations substantially improves robustness and generalization for open-world robotic manipulation.
Project website: \href{https://oucx117.github.io/ST-VLA/}{https://oucx117.github.io/ST-VLA/}.
\end{abstract}

%% file: sections/2_intro.tex
\section{Introduction}
\label{sec:intro}

Robust robot manipulation in open-world environments requires tightly integrating generalizable semantic understanding, accurate spatial perception, and reliable execution. Beyond identifying interaction targets, robots must reason about \emph{how and where} to act in complex three-dimensional (3D) environments. Hierarchical Vision-Language-Action (VLA) architectures have therefore emerged as a dominant paradigm, where high-level Vision-Language Models (VLMs) perform task understanding and semantic reasoning, and low-level policies execute continuous control, improving generalization across manipulation tasks~\cite{brohan2022rt,zitkovich2023rt,driess2023palm,kim2024openvla,ahn2022can,reed2022generalist,chen2026deco}.

Despite this progress, existing VLMs are primarily trained in large-scale two-dimensional (2D) images and visual question answering data, limiting their ability to model 3D geometry and long-horizon manipulation dynamics in physical environments~\cite{radford2021learning,liu2023visual}. As a result, high-level semantic predictions often fail to align with the continuous 3D execution space required for fine-grained robotic control, leading to deviations from action, instability, and task failure~\cite{ahn2022can,huang2023voxposer}.
To alleviate this issue, previous work introduces intermediate representations such as 2D waypoints, bounding boxes, or segmentation masks to connect semantic reasoning with low-level control~\cite{liu2024moka,gu2023rt,ren2024grounded,jiang2022vima,bharadhwaj2024track2act,yuan2024robopoint,shridhar2022cliport}. However, these interfaces remain fundamentally constrained by the 2D paradigm. Planar representations cannot capture depth and geometric constraints, and frame-wise signals ignore temporal continuity~\cite{ke20243d,huang2023voxposer,chi2025diffusion,liu2022structdiffusion,goyal2023rvt}. Consequently, they lack explicit modeling of 4D (3D + time) spatio-temporal consistency, often inducing action jitter and reduced robustness in dynamic or long-horizon tasks~\cite{zheng2024tracevla}.

We argue that these limitations stem from a fundamental \emph{representation mismatch} between high-level semantic spaces and low-level physical spaces~\cite{driess2023palm,huang2023voxposer,kerr2023lerf,peng2023openscene,jatavallabhula2023conceptfusion}. VLM representations are rooted in static, projective visual domains, whereas robot control operates in continuous 3D physical environments. As shown in Figure~\ref{fig:teaser}(a),the 2D intermediates form a lossy bridge that discards critical geometric and temporal cues. This forces low-level policies to infer missing dimensions from ambiguous signals, leading to spatial misalignment and execution inaccuracy under novel viewpoints or disturbances~\cite{huang2023voxposer,ke20243d}, as well as temporal incoherence that hinders long-horizon stability~\cite{chen2024scar,chen2025robohorizon}. An effective intermediate representation should instead jointly capture 3D geometry, temporal continuity, and cross-modal consistency.

Motivated by this insight, we propose \textbf{\method}, a hierarchical VLA framework that bridges the perception-action gap via unified 3D-4D representations. As shown in Figure~\ref{fig:teaser}(b), instead of rigid 2D cropping, \method lifts conventional 2D guidance to explicit 3D trajectories and constructs cross-modally aligned smooth spatial masks to inpaint task-irrelevant distractors. This smooth alignment ensures focus on task-relevant geometry, which preserves latent continuity, suppresses model hallucinations, and improves action stability. Moreover, by integrating 4D spatio-temporal context, \method enables online replanning and robust long-horizon execution. 
To enable high-level VLMs to produce 3D-4D representations, we introduce \textbf{ST-Human}, a large-scale dataset with 4.3M samples, built from 300K continuous human manipulation trajectories paired with high-fidelity 3D observations. Unlike prior 2D-centric or static datasets, ST-Human captures depth, object geometry, and temporal action evolution, providing unified spatio-temporal supervision. 
Building on ST-Human, we propose a unified 2D-3D-4D task learning framework that extends VLM semantic reasoning across 3D spatial and 4D temporal domains. By jointly fine-tuning a 4B Qwen3-VL model~\cite{yang2025qwen3} on ST-Human alongside specialized public datasets—including RoboPoint~\cite{yuan2024robopoint}, FSD~\cite{yuan2025seeing}, and SAT~\cite{ray2024sat}—we develop ST-VLM. This unified training approach integrates 2D trajectory grounding, depth-aware 3D perception, and long-horizon 4D reasoning, enabling the model to acquire comprehensive spatio-temporal understanding within a single representative space. With explicit geometric and temporal constraints, ST-VLM learns stable, transferable 3D--4D reasoning, enabling reliable zero-shot guidance for open-world manipulation.

We evaluate ST-VLM and ST-VLA on multiple benchmark datasets as well as simulated and real-world robotic platforms. On the RoboRefit~\cite{lu2023vl}, CVBench~\cite{tong2024cambrian}, and SAT~\cite{ray2024sat} datasets, ST-VLM achieves up to 33.19\% improvement over existing methods. On 3 representative RLBench tasks, ST-VLA improves zero-shot success rates by 44.6\%. Real-world long-horizon experiments further demonstrate its execution stability and cross-scenario generalization.

Our contributions are summarized as follows: \textbf{(1)} We propose \textbf{\method}, a hierarchical Vision-Language-Action (VLA) framework built upon unified 3D-4D (3D + time) intermediate representations.
\method replaces conventional, error-prone 2D priors with explicit 3D trajectories and latent 4D spatio-temporal context, ensuring action stability and suppressing hallucinations. \textbf{(2)} We introduce \textbf{ST-Human}, a high-fidelity 3D-4D human manipulation dataset,and propose a unified 2D-3D-4D task learning and evaluation framework. Based on this framework, we train a 4B-parameter vision-language model, \textbf{ST-VLM}, which acquires transferable 3D-4D reasoning capabilities for zero-shot open-world manipulation. \textbf{(3)} We evaluate ST-VLM and ST-VLA on multiple benchmarks and simulated and real-world robotic platforms. Results show clear gains in zero-shot success rate, long-horizon stability, and cross-scenario generalization over existing methods.

%% file: sections/3_related_work.tex
\begin{figure*}[t] %
    \centering
    \includegraphics[width=\linewidth]{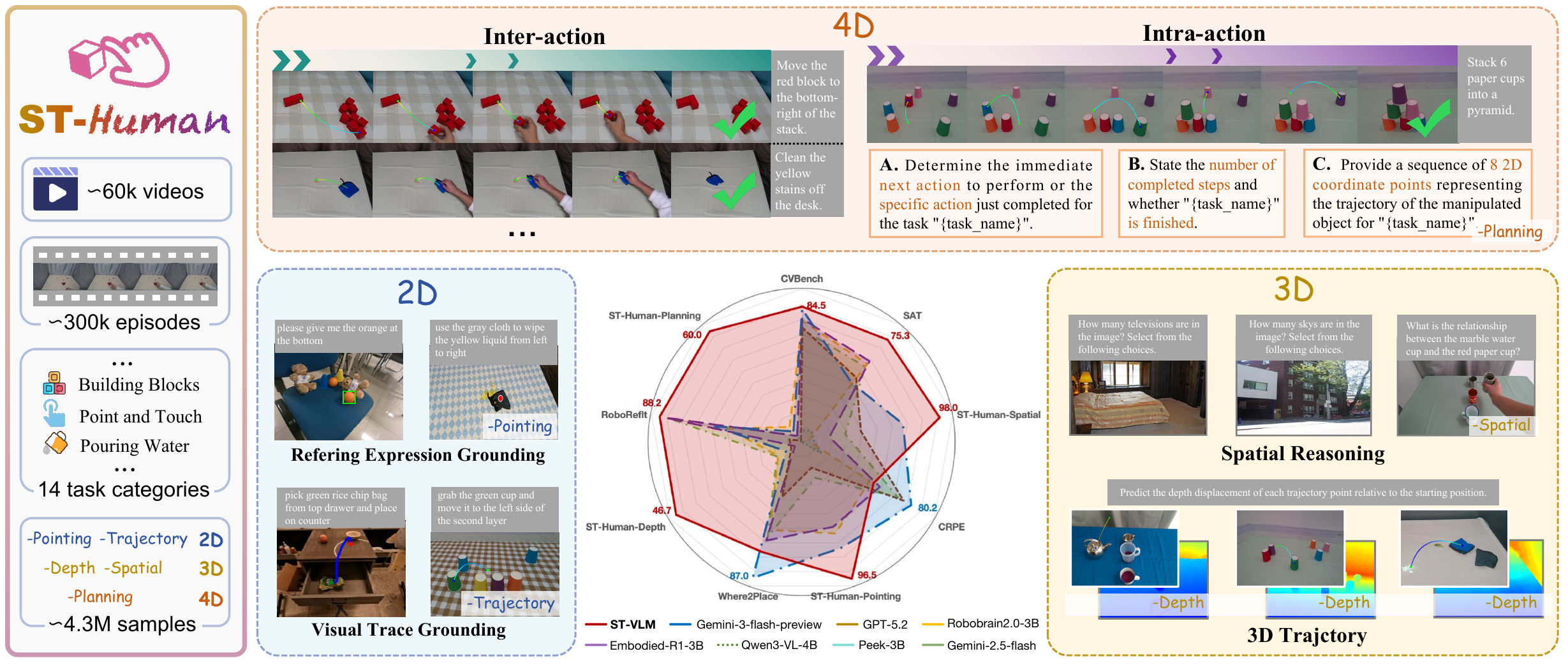} 
    \caption{Overview of the ST-Human Dataset Construction and Unified 2D-3D-4D Task Generation.}
    \label{fig:vlm}
    % \vspace{-5mm}
\end{figure*}

\section{Related Work}

\subsection{Point and Trajectory-based Spatial Guidance}
An important line of research in manipulation focuses on the generalization of robotic policies by providing explicit spatial guidance. Early works utilize 2D waypoints or bounding boxes \cite{liu2024moka,huang2024manipvqa} to anchor the policy’s attention to specific image-plane coordinates. 
Beyond image-plane cues, another line explores 3D value-based spatial representations as guidance for manipulation~\cite{chengravmad,huang2023voxposer}. 
More recently, trajectory-based task specifications emerge as a flexible interface. For example, RT-Trajectory \cite{gu2023rt} and Track2Act \cite{bharadhwaj2024track2act} employ 2D gripper paths or point tracks to condition low-level execution. HAMSTER \cite{li2025hamster} further extends this paradigm hierarchically, where a VLM predicts future 2D paths to guide a conditioned 3D policy. However, these representations remain fundamentally restricted to the projective image domain, lacking the geometric fidelity and depth awareness required for precision-critical 3D manipulation. Moreover, since such signals are typically generated on a per-frame basis, they often fail to capture the temporal coherence \cite{zheng2024tracevla} necessary for dynamic replanning.

\subsection{Object-Centric Grounding and Masking}
To mitigate the impact of visual distractors in open-world environments, recent methods have turned to object-centric representations \cite{shi2024plug, mirjalili2025augmented, hancock2025run, yuan2025roboengine, huang2025otter, li2025controlvla}. Frameworks such as ARRO \cite{mirjalili2025augmented} and PEEK \cite{zhang2025peek} leverage pre-trained segmentation models or VLM-queried points to visually isolate task-relevant objects through explicit masking. While effective at filtering clutter, existing approaches predominantly utilize hard segmentation masks like cropping in RoboPoint\cite{yuan2024robopoint}, which, as identified in recent reasoning-driven studies \cite{zhao2025cot,yuan2025seeing}, can introduce discontinuities in the latent manifold. These discrete representations often lead to jittery action outputs and hallucinations due to cross-modal misalignment.

\subsection{Vision-Language Reasoning for Manipulation}
Robotic manipulation has progressed from using static vision-language representations \cite{parisi2022unsurprising,nair2022r3m} to end-to-end Vision-Language-Action (VLA) models \cite{brohan2024rt,kim2024openvla}. Although foundation VLMs provide broad open-world semantics, their physical reasoning and generalization are often limited by weak embodied grounding \cite{brohan2022rt}. To mitigate this, recent work improves spatial intelligence via supervised fine-tuning on spatial datasets \cite{chen2024spatialvlm} or uses reinforcement learning to elicit stronger reasoning \cite{guo2025deepseek,yuan2025embodied}.
A key bottleneck is the lack of high-quality continuous 3D/4D data linking semantic intent to execution. Existing VLA systems largely rely on large-scale robot demonstrations \cite{walke2023bridgedata,khazatsky2024droid}, which are costly to scale, or simulation-to-real transfer \cite{james2020rlbench,tao2024maniskill3}, which remains hindered by domain gaps. In contrast, we propose ST-Human, a high-fidelity human manipulation dataset with dense 3D/4D annotations that capture continuous execution flows, enabling ST-VLM to learn anchor-based depth estimation and long-horizon reasoning.

%% file: sections/4_method.tex
\section{Methodology}

\subsection{Problem Formulation}

\begin{figure*}[t] %
    \centering
    \includegraphics[width=\linewidth]{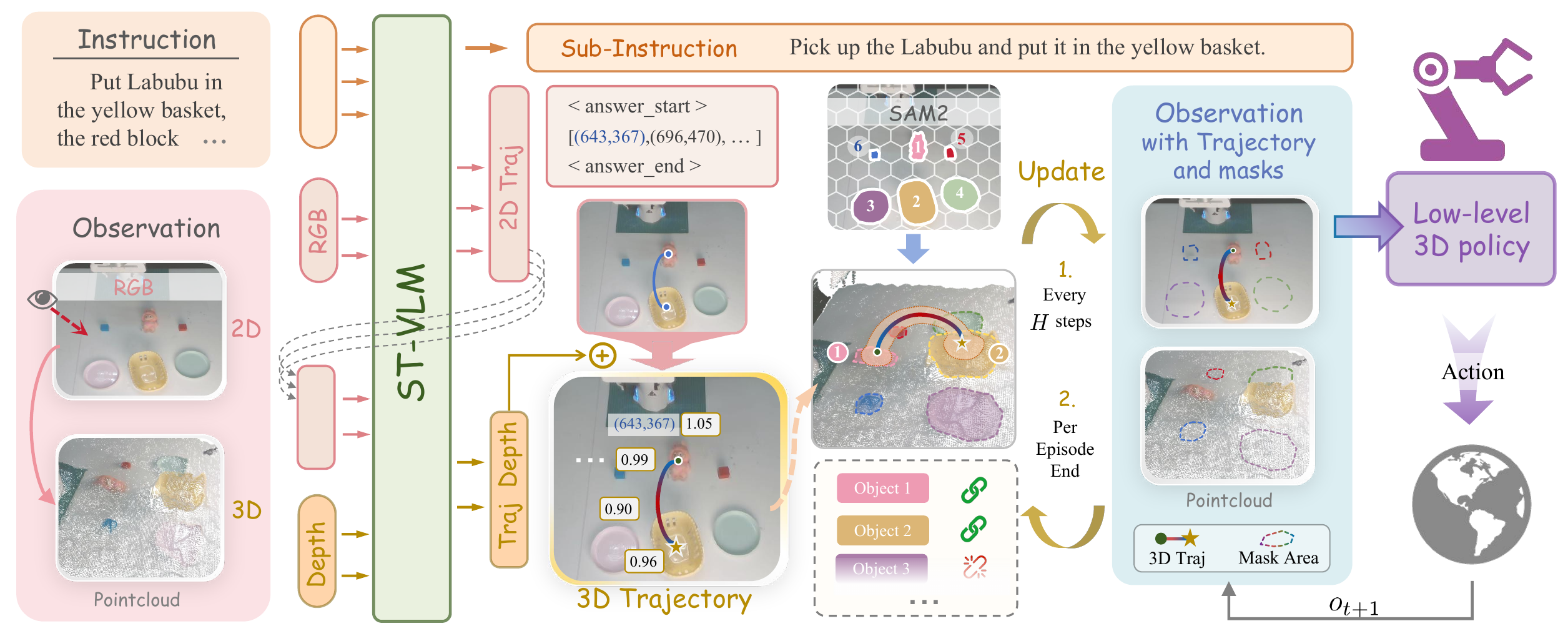} 
    \caption{\textbf{The ST-VLA Pipeline.} Given a global instruction and an RGB-D observation, the high-level \textbf{ST-VLM} generates sub-instructions and 2D trajectories. These are lifted to 3D and fused with SAM2 masks to form a unified 3D-4D representation, which conditions the low-level 3D policy for continuous action execution. Guidance is refreshed every $H$ steps for replanning and robustness to disturbances.}
    \label{fig:pipeline}
    % \vspace{-5mm}
\end{figure*}

We consider the problem of an embodied agent executing manipulation tasks in open-world environments guided by a natural language instruction $l \in \mathcal{L}$. At each timestep $t$, the agent receives a single-view RGB-D observation $\mathbf{o}_t \in \mathcal{O}$ and a robot state $\mathbf{s}_t \in \mathcal{S}$. Following the established paradigm of 3D-aware policies, the raw observation $\mathbf{o}_t$ is unprojected into a 3D point cloud $\mathbf{P}_t$ in the robot’s base frame. The objective is to learn a mapping that generates a sequence of actionable keyposes $\mathbf{a}_t \in SE(3)$, which are subsequently translated by a deterministic motion planner into continuous joint control commands.

\subsection{Hierarchical VLA Architecture Overview} 
\label{sec:hvla}
To bridge semantic reasoning with physical execution, we decouple the manipulation policy into a hierarchy: a high-level vision-language model $\pi_{hi}$ and a low-level execution policy $\pi_{lo}$. 
The high-level $\pi_{hi}$ is derived from a pre-trained general-purpose VLM and is specialized through supervised fine-tuning on large-scale datasets encompassing diverse robotic manipulation tasks. This specialization enables the VLM to function as a spatio-temporal reasoner. Given the observation $\mathbf{o}_t$ and instruction $l$, $\pi_{hi}$ generates a unified intermediate representation $\mathcal{Z}$ and, in long-horizon scenarios, produces an updated sub-instruction $l'$ to guide subsequent execution. This process is formulated as:
\begin{equation}
(\mathcal{Z}, l') = \pi_{hi}(\mathbf{o}_t, l),
\end{equation}
where $\mathcal{Z} = \{\tau, \mathcal{M}\}$ consists of a 3D trajectory $\tau$ and a corresponding spatial mask $\mathcal{M}$. In short-horizon tasks where the initial instruction is sufficient, the instruction remains invariant such that $l' = l$.

The low-level policy $\pi_{lo}$ is responsible for control within an augmented observation space. We define an augmentation function $\psi$ that integrates the spatio-temporal guidance $\mathcal{Z}$ into the observation stream $\tilde{\mathbf{o}}_t = \psi(\mathbf{o}_t, \mathcal{Z})$. This allows $\pi_{lo}$ to use the spatial priors from $\pi_{hi}$ without architectural changes. The spatial masks in $\mathcal{M}$ are implemented with smooth boundaries for stable feature extraction. Finally, $\pi_{lo}$ maps the augmented observation, the robot state, and the potentially updated instruction to the target keypose $\mathbf{a}_t$:
\begin{equation}
\mathbf{a}_t = \pi_{lo}(\tilde{\mathbf{o}}_t, \mathbf{s}_t, l').
\end{equation}
By offloading semantic reasoning and spatial grounding to $\pi_{hi}$, our framework enables $\pi_{lo}$ to focus on geometric execution and enhance zero-shot generalization.

\subsection{Spatio-Temporal VLM Learning with ST-Human}
To support the hierarchical architecture in Sec.~\ref{sec:hvla}, we introduce \textbf{ST-Human}, a high-fidelity dataset that provides the supervision needed for a general-purpose VLM $\pi_{hi}$ to learn spatial and temporal reasoning. This section describes large-scale data collection and the generation of 2D, 3D, and 4D supervisory tasks.

\subsubsection{Scalable 3D-4D Dataset Construction}

We record \(\sim\)60k videos (300k episodes) across 14 single-arm tabletop manipulation categories using a fixed RGB-D sensor; see Appendix~\ref{app:task-suite}. To ground the pipeline in physical constraints, we annotate initial/final 2D contact points, object semantics, and relative spatial configurations for a core subset, and develop an annotation interface (Appendix~\ref{app:annotation-framework}). These labels serve as deterministic seeds for automated annotation at scale, reducing error accumulation compared to purely heuristic methods.

Following collection, a fully automated pipeline organizes supervision into 2D, 3D, and 4D task groups for multi-task fine-tuning (Figure~\ref{fig:vlm}). \textbf{2D spatial tasks} target image-plane grounding and tracking. We back-project the annotated 3D grasp points to pixel coordinates for semantic-to-pixel alignment and utilize SAM2 with video tracking to generate 2D keypoint trajectories. \textbf{3D geometric tasks} incorporate depth and relational structure. We synthesize 3D trajectories by fusing 2D motion flows with aligned depth maps and lifting waypoints into workspace coordinates. A relational module converts spatial labels into dense supervision over relative object positions and orientations, forming a scene-graph style representation. Finally, \textbf{4D spatio-temporal tasks} model action evolution over time. By linking dependent episodes within the same video, we train $\pi_{hi}$ to reason about task progress and plan across actions. Within each episode, $\pi_{hi}$ predict the remaining trajectory from partial observations, encouraging dynamic replanning and maintaining representation coherence during execution. See Appendix~\ref{app:algorithms} for details.

%------------------------------------------------------------------------------------------

\definecolor{bestblue}{RGB}{235, 242, 255} 
\definecolor{groupgray}{RGB}{247, 247, 247} 

\begin{table*}[t]
\centering
\small
\caption{\textbf{VLM Performance Comparison on Selected 2D, 3D, and 4D Benchmarks.} Best results are bolded.}
\label{tab:main_results}
\setlength{\tabcolsep}{3.5pt} 
\renewcommand{\arraystretch}{1.1} 

\resizebox{\textwidth}{!}{%
\begin{tabular}{l ccc ccccc c}
\toprule
\multirow{2}{*}{\textbf{Method}} & \multicolumn{3}{c}{\textbf{2D Tasks}} & \multicolumn{5}{c}{\textbf{3D Tasks}} & \textbf{4D Tasks} \\
\cmidrule(lr){2-4} \cmidrule(lr){5-9} \cmidrule(lr){10-10}
& RoboRefit & Where2Place & ST-Human-Pointing & CVBench & SAT & CRPE & ST-Human-Spatial & ST-Human-Depth & ST-Human-Planning \\
\midrule

% Closed-source Section
\rowcolor{groupgray} \multicolumn{10}{l}{\textit{Closed-source models}} \\
GPT-5.2         & 16.65\% & 43.00\% & 20.50\% & 79.62\% & 66.00\% & 78.89\% & 68.00\% & 4.00\%  & 88.00\% \\
Gemini-3-flash  & 57.75\% & \textbf{87.00\%} & 75.50\% & 83.94\% & 64.67\% & \textbf{80.21\%} & 84.00\% & 6.00\%  & 84.00\% \\

% Open-source Section
\rowcolor{groupgray} \multicolumn{10}{l}{\textit{Open-source models}} \\
PEEK-3B         & 10.05\% & 14.00\% & 3.00\%  & 55.49\% & 52.67\% & 64.55\% & 50.00\% & 0.00\%  & 88.00\% \\
Qwen3VL-4B      & 83.63\% & 65.00\% & 27.50\% & 79.21\% & 68.67\% & 77.89\% & 62.00\% & 9.33\%  & 88.00\% \\
Robobrain2.0-3B & 50.61\% & 61.00\% & 65.50\% & 81.22\% & 69.33\% & 73.03\% & 48.00\% & 8.89\%  & 88.00\% \\
Embodied-R1-3B  & 85.05\% & 66.83\% & 61.00\% & 81.69\% & 70.00\% & 74.78\% & 56.00\% & 2.70\%  & 78.00\% \\

% Ours Section
\midrule 
\rowcolor{bestblue} 
\textbf{ST-VLA (Ours)} & \textbf{88.15\%} & 73.00\% & \textbf{96.50\%} & \textbf{84.52\%} & \textbf{75.33\%} & 73.67\% & \textbf{98.00\%} & \textbf{46.67\%} & \textbf{92.00\%} \\
\bottomrule
\end{tabular}%
}
\end{table*}
%------------------------------------------------------------------------------------------

\subsubsection{Unified VLM Fine-tuning and Spatio-Temporal Inference}

To equip $\pi_{hi}$ with specialized 3D-4D reasoning, we implement a two-stage supervised fine-tuning (SFT) strategy and a hierarchical inference protocol that lifts 2D visual cues to 3D execution space (see Figure ~\ref{fig:pipeline}).

During training, we first leverage public multimodal datasets for general semantic reasoning, followed by domain-specific SFT on ST-Human. 
The prompts used for fine-tuning are provided in Appendix~\ref{app:prompts-format}.
The model learns to map instructions and observations to structured text outputs, including normalized 2D coordinates \(\mathbf{x} \in [0,1]^2\). 
Building on these representations, we resolve the geometric ambiguity that a planar trajectory \(\tau_{2D}\) may map to multiple 3D paths. Given an RGB image and a colorized depth map, \(\pi_{hi}\) predicts \(\tau_{2D}\) and retrieves the ground-truth starting depth \(d_{start}\) from the depth input. It then predicts relative depth offsets for subsequent waypoints. These offsets represent intended end-effector depth rather than static surface depth.

During deployment, the VLM runs every \(H\) timesteps, balancing compute efficiency with stable long-horizon planning. Each update follows two stages.
First, the VLM predicts a projected 2D trajectory \(\tau_{2D}\) from the target object to the goal given the current RGB observation and instruction \(l\). Second, it lifts \(\tau_{2D}\) to 3D  \(\tau_{3D}\) by combining RGB-D observations with the anchored starting depth \(d_{start}\) and predicting relative depth offsets along the waypoints. The resulting \(\tau_{3D}\) provides a continuous spatial condition for the low-level policy \(\pi_{lo}\) until the next update.

\begin{table*}[t]
\centering
\small
\caption{\textbf{Results on Simulated Robot Manipulation Tasks.} Background colors highlight ST-VLA variants.}
\label{tab:results}
\setlength{\tabcolsep}{12pt}
\renewcommand{\arraystretch}{0.5}
\resizebox{\textwidth}{!}{
\begin{tabular}{l c cc cc cc cc}
\toprule
\multirow{2}{*}{\textbf{}} & \multirow{2}{*}{\textbf{Method}} & \multicolumn{2}{c}{\textbf{Close Jar}} & \multicolumn{2}{c}{\textbf{Light Bulb In}} & \multicolumn{2}{c}{\textbf{Put Groceries}} & \multicolumn{2}{c}{\textbf{Average}} \\
\cmidrule(lr){3-4} \cmidrule(lr){5-6} \cmidrule(lr){7-8} \cmidrule(lr){9-10}
& & Seen & Unseen & Seen & Unseen & Seen & Unseen & Seen & Unseen \\
\midrule
\multirow{7}{*}{\rotatebox{90}{\textbf{Single-task}}}
& 3DDA (Baseline) \rule{0pt}{2ex} & 71.9\std{3.0} & 44.4\std{9.6} & 20.8\std{3.6} & 25.9\std{6.4} & 33.3\std{3.0} & 00.0\std{0.0} & 42.0 & 23.4 \\
& \cellcolor{gray!10} ST-VLA w/ 3DDA (Frozen) \rule{0pt}{2ex} & \cellcolor{gray!10}98.2\std{3.0} & \cellcolor{gray!10}66.7\std{0.0} & \cellcolor{gray!10}37.5\std{0.0} & \cellcolor{gray!10}25.9\std{17.0} & \cellcolor{gray!10}42.1\std{5.3} & \cellcolor{gray!10}5.6\std{9.6} & \cellcolor{gray!10}59.3 & \cellcolor{gray!10}32.7\\
& \cellcolor{gray!22} \textbf{ST-VLA w/ 3DDA (FT)} \rule{0pt}{2ex} & \cellcolor{gray!22}\textbf{100.0}\std{0.0} & \cellcolor{gray!22}\textbf{100.0}\std{0.0} & \cellcolor{gray!22}\textbf{45.8}\std{3.6} & \cellcolor{gray!22}\textbf{81.5}\std{6.4} & \cellcolor{gray!22}\textbf{54.4}\std{8.0} & \cellcolor{gray!22}\textbf{16.7}\std{16.7} & \cellcolor{gray!22}\textbf{66.7} & \cellcolor{gray!22}\textbf{66} \\
\addlinespace[4pt]
& 3DFA (Baseline) \rule{0pt}{2ex} & 61.4\std{3.0} & 38.9\std{9.6} & 22.9\std{3.6} & 11.1\std{0.0} & 5.3\std{5.3} & 00.0\std{0.0} & 29.9 & 16.7 \\
& \cellcolor{blue!5} ST-VLA w/ 3DFA (Frozen) \rule{0pt}{2ex} & \cellcolor{blue!5}68.4\std{5.3} & \cellcolor{blue!5}44.4\std{9.6} & \cellcolor{blue!5}\textbf{41.7}\std{3.6} & \cellcolor{blue!5}11.1\std{0.0} & \cellcolor{blue!5}12.3\std{3.0} & \cellcolor{blue!5}00.0\std{0.0} & \cellcolor{blue!5}40.8 & \cellcolor{blue!5}18.5\\
& \cellcolor{blue!11} \textbf{ST-VLA w/ 3DFA (FT)} \rule{0pt}{2ex} & \cellcolor{blue!9}\textbf{91.2}\std{6.1} & \cellcolor{blue!9}\textbf{72.2}\std{9.6} & \cellcolor{blue!9}{39.6}\std{7.2} & \cellcolor{blue!9}\textbf{59.3}\std{6.4} & \cellcolor{blue!9}\textbf{56.1}\std{8.0} & \cellcolor{blue!9}\textbf{11.1}\std{9.6} & \cellcolor{blue!9}\textbf{62.3} & \cellcolor{blue!9}\textbf{47.5} \\
\specialrule{0.6pt}{3pt}{3pt} 
\multirow{7}{*}{\rotatebox{90}{\textbf{Multi-task}}} 
& 3DDA (Baseline) \rule{0pt}{2ex} & 75.4\std{3.0} & 11.1\std{9.6} & 56.3\std{0.0} & 37.0\std{12.8} & 63.2\std{5.3} & 00.0\std{0.0} & 65.0 & 16.0 \\
& \cellcolor{gray!10} ST-VLA w/ 3DDA (Frozen) \rule{0pt}{2ex}  & \cellcolor{gray!10}\textbf{100.0}\std{0.0} & \cellcolor{gray!10}55.6\std{9.6} & \cellcolor{gray!10}\textbf{68.8}\std{6.3} & \cellcolor{gray!10}59.3\std{6.4} & \cellcolor{gray!10}54.4\std{3.0} & \cellcolor{gray!10}00.0\std{0.0} & \cellcolor{gray!10}74.4& \cellcolor{gray!10}38.3 \\
& \cellcolor{gray!22} \textbf{ST-VLA w/ 3DDA (FT)} \rule{0pt}{2ex} & \cellcolor{gray!22}{98.2}\std{3.0} & \cellcolor{gray!22}\textbf{100.0}\std{0.0} & \cellcolor{gray!22}{54.2}\std{3.6} & \cellcolor{gray!22}\textbf{81.5}\std{6.4} & \cellcolor{gray!22}\textbf{73.7}\std{0.0} & \cellcolor{gray!22}\textbf{22.2}\std{9.6} & \cellcolor{gray!22}\textbf{75.4} & \cellcolor{gray!22}\textbf{67.9} \\
\addlinespace[4pt]
& 3DFA (Baseline) \rule{0pt}{2ex} & 71.9\std{3.0} & 44.4\std{9.6} & \textbf{66.7}\std{7.2} & 00.0\std{0.0} & 49.1\std{3.0} & 5.6\std{9.6} & 62.6 & 16.7 \\
& \cellcolor{blue!5} ST-VLA w/ 3DFA (Frozen) \rule{0pt}{2ex} & \cellcolor{blue!5}93.3\std{3.0} & \cellcolor{blue!5}67.7\std{16.7} & \cellcolor{blue!5}{64.6}\std{3.6} & \cellcolor{blue!5}29.6\std{6.4} & \cellcolor{blue!5}56.1\std{11.0} & \cellcolor{blue!5}11.1\std{9.6} & \cellcolor{blue!5}\textbf{71.3} & \cellcolor{blue!5}36.1 \\
& \cellcolor{blue!9} \textbf{ST-VLA w/ 3DFA (FT)} \rule{0pt}{2ex} & \cellcolor{blue!9}\textbf{98.2}\std{3.0} & \cellcolor{blue!9}\textbf{100.0}\std{0.0} & \cellcolor{blue!9}{54.2}\std{3.6} & \cellcolor{blue!9}\textbf{81.5}\std{6.4} & \cellcolor{blue!9}\textbf{56.1}\std{3.0} & \cellcolor{blue!9}\textbf{27.8}\std{9.6} & \cellcolor{blue!9}{69.5} & \cellcolor{blue!9}\textbf{69.7}\\
\bottomrule
\end{tabular}
}
\end{table*}

\begin{table}[h]
\centering
\small
\caption{Success rates on long-horizon \textit{push-button} tasks across three seeds, evaluated ST-VLA(3DFA) on seen buttons and unseen multi-step sequences.}
\label{tab:pushbutton_results}
\setlength{\tabcolsep}{15pt} 
\renewcommand{\arraystretch}{0.6}
\begin{tabular}{lc}
\toprule
\textbf{Category} & \textbf{Success Rate (\%)} \\
\midrule
Seen Tasks & 97.4\std{4.4} \\
LH 2-step (Unseen) & 93.3\std{11.5} \\
LH 3-step (Unseen) & 100.0\std{0.0} \\
\midrule
\textbf{Overall} & \textbf{97.3}\std{2.3} \\
\bottomrule
\end{tabular}
\vspace{-5mm}
\end{table}

\subsection{Hierarchical 3D Control via Unified Representation}

To bridge reasoning and execution, we define a unified intermediate representation \(\mathcal{Z}\) that combines spatial-temporal paths and geometric constraints into a single guidance signal. This section describes how \(\mathcal{Z}\) is generated and how it is injected cross-modally to condition the low-level policy.

\subsubsection{Spatio-Temporal Guidance Construction}
\label{sec:st_guidance}

In long-horizon scenarios, the high-level model $\pi_{hi}$ first decomposes the initial global instruction $l$ into atomic, action-specific sub-instructions $l'$, whereas $l'$ and $l$ are equivalent in short-term tasks. Guided by the refined instruction, $\pi_{hi}$ predicts the 3D trajectory $\tau_{3D} = \{\mathbf{p}_1, \dots, \mathbf{p}_K\}$, which is later expanded into a spatial tube $\mathcal{T}$. This tube is defined by a cylindrical expansion with a fixed radius $r$, such that $\mathcal{T} = \bigcup_{k=1}^K \mathcal{B}(\mathbf{p}_k, r)$, where $\mathcal{B}(\mathbf{p}, r)$ denotes a 3D ball centered at $\mathbf{p}$. $\mathcal{T}$ highlights the safe and intended operational volume for the robot's end-effector.

We introduce a selective masking mechanism to isolate task-relevant geometry.
First, we unproject SAM2-generated dense instance segmentation $\{M_i\}$ into the 3D workspace. An object is identified as task-relevant if its occupancy $\mathcal{V}_i$ intersects with the spatial tube, i.e., $\mathcal{V}_i \cap \mathcal{T} \neq \emptyset$. 
For all task-irrelevant objects, we apply a cross-modal smoothing operator to both RGB and depth channels. Non-relevant mask regions are inpainted via interpolation, constrained by RGB color gradients and neighboring depth values to preserve local feature coherence. This approach ensures cross-modal alignment and prevents instability in the policy's latent manifold. 

The final augmented observation $\tilde{\mathbf{o}}_t$ overlays the 2D projection of $\tau_{3D}$ on the inpainted RGB image with a translucent red-to-blue gradient to encode depth. Together with the refined instruction $l'$, it is fed to the low-level policy $\pi_{lo}$, focusing attention on the task-relevant spatial manifold.

\subsubsection{3D-Aware Policy Specialization}

We adopt 3D Diffuser Actor (3DDA) \cite{ke20243d} and 3D FlowMatch Actor (3DFA) \cite{gkanatsios20253d} as low-level backbones for their robust 3D-aware representations from single-view RGB-D input. 3DDA models action distributions via denoising, while 3DFA regresses trajectory segments through flow matching. Both map the unprojected point cloud $\mathbf{P}_t$ to accurate $SE(3)$ keyposes for execution by a motion planner. We specialize these models via imitation learning on augmented expert demonstrations. Rather than using dense trajectory policies, we adopt keypose-based training, extracting sequential keypose coordinates to reconstruct the ground-truth 3D guidence.

During training, we apply the same augmentation function $\psi$ (Sec.~\ref{sec:st_guidance}) to the raw observations. Concretely, we render the ground-truth trajectory onto the RGB image and apply cross-modal smooth spatial masks to the RGB-D input. This strategy encourages the policy $\pi_{lo}$ to align its attention with the VLM-provided spatio-temporal priors. At inference time $\pi_{lo}$ can reliably track the operation tube while suppressing distractors outside the masked regions.

A key advantage of our smooth masking is its robustness to observation shifts. Unlike traditional brittle binary truncation, smooth boundary transitions preserve feature continuity and manifold stability.
This design further allows our framework to function with vanilla policies trained on raw, unaugmented data. In such frozen-policy configurations, where the low-level policy $\pi_{lo}$ is used without task-specific retraining on augmented data. The smooth mask acts as a high-level attention filter that concentrates the policy's focus on task-relevant geometries. We omit the explicit trajectory visualization to avoid confusing the frozen policy, retaining only the spatial masking. 

%% file: sections/5_exp.tex
\section{Experiments}

\subsection{Evaluation of ST-VLM Capabilities}

We evaluate our model on nine benchmarks across 2D, 3D, and 4D dimensions (Table \ref{tab:main_results}). This suite includes established open-source datasets (RoboRefit~\cite{lu2023vl}, Where2Place~\cite{yuan2024robopoint}, CVBench~\cite{tong2024cambrian}, SAT~\cite{ray2024sat}, and CRPE~\cite{wang2024all}) and our proprietary ST-Human series, which comprises ST-Human-Pointing (Point Detection), -Trajectory (Predicting 2D trajectory), -Spatial (Spatial Understanding), -Depth (Trajectory Depth Estimation), and -Planning (4D Temporal Reasoning).
\begin{figure*}[t] %
    \centering
    \includegraphics[width=\linewidth]{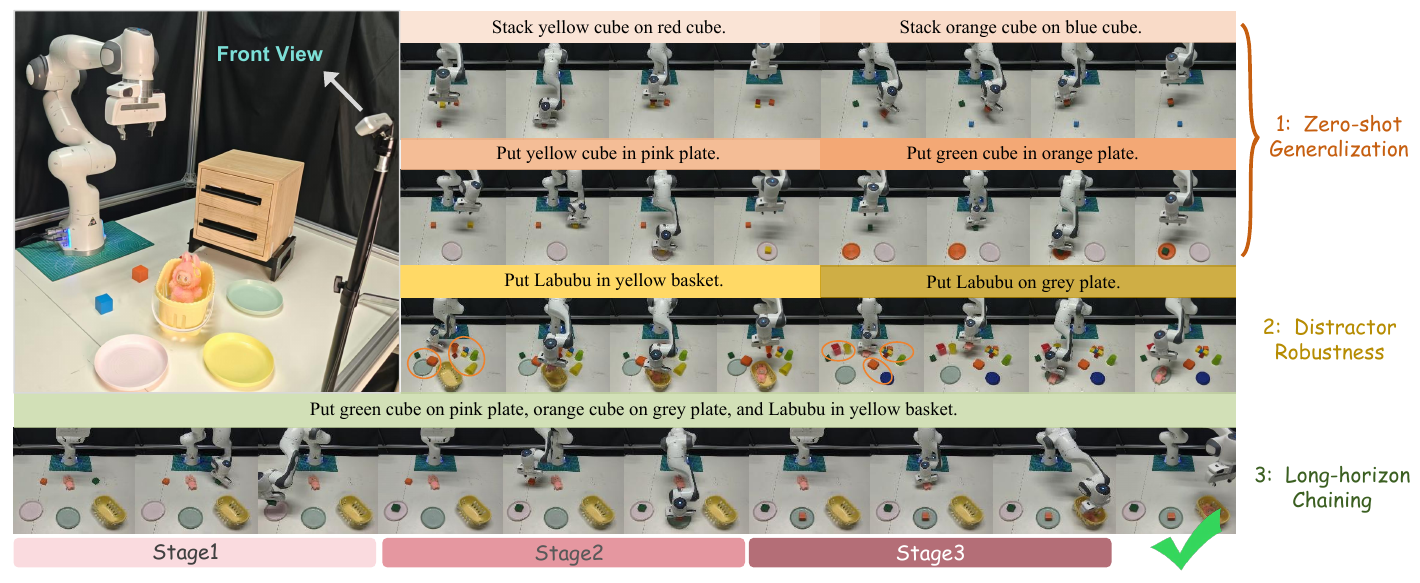} 
    \caption{\textbf{Real-world experimental evaluation of ST-VLA}. Left: Franka Emika Panda hardware setup. Right: Qualitative results across three dimensions: (1) zero-shot generalization to unseen object categories and geometries; (2) distractor robustness under task-irrelevant visual clutter; and (3) long-horizon chaining via sequential execution of multiple placement tasks.}
    \label{fig:real_env_setup}
    
\end{figure*}

\begin{figure}[t] %
    \centering
    \includegraphics[width=\linewidth]{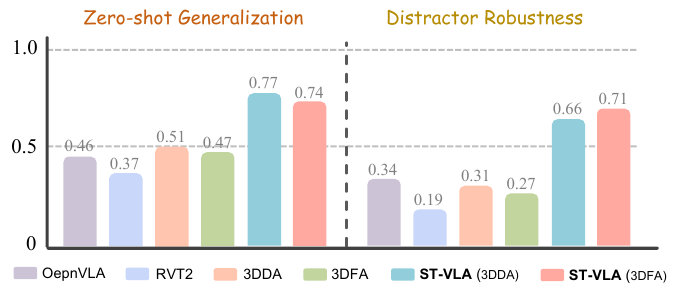} 
    \caption{Real-world zero-shot generalization results.}
    \label{fig:real_world_results}
    
\end{figure}
We compare our model against six state-of-the-art (SOTA) baselines, consisting of four specialized open-source models: Embodied-R1-3B~\cite{Embodied-r1}, PEEK-3B~\cite{zhang2025peek}, Robobrain2.0-3B~\cite{team2025robobrain}, and our base model Qwen3VL-4B~\cite{yang2025qwen3}; and two leading closed-source models: GPT-5.2~\cite{openai2025gpt52} and Gemini-3-pro-preview~\cite{google2025gemini3propreview}. All evaluations are conducted across both existing open-source benchmarks and our curated \methodset dataset to assess zero-shot generalization and task-specific precision. 
Detailed hyperparameters are provided in Table~\ref{tab:qwen3_vlsft_params}. More experimental results on VLM capabilities and metric definitions are provided in Appendix~\ref{sec:appendix_results}.

The comparative results are summarized in Table~\ref{tab:main_results}. Overall, our model achieves competitive performance across all categories. Several key observations emerge from the data:

\textit{1) Superior Spatio-Temporal Precision:} ST-VLA demonstrates a significant lead in ST-Human-Depth and 4D Tasks. It achieves 46.67\% accuracy in depth estimation, far surpassing the best baseline (9.33\%), and maintains temporal coherence in long-horizon planning (92.00\% \textit{vs.} 88.00\%). This performance gain is attributed to the high-fidelity annotations in \methodset, which empower the VLM to transition from static 2D grounding to dynamic 3D-aware reasoning.

\textit{2) Robustness in 3D Understanding:} While closed-source models like GPT-5.2 show strong semantic reasoning (e.g., on CVBench), they often struggle with precise spatial grounding in robotic workspaces. ST-VLA outperforms these general-purpose VLMs on ST-Human-Spatial (98.00\% \textit{vs.} 84.00\%) and ST-Human-Pointing (96.50\% \textit{vs.} 75.50\%), indicating that specialized fine-tuning on embodied data is essential for manipulation-level precision.

\textit{3) Efficiency and Generalization:} Despite having fewer parameters than largest closed-source counterparts like Gemini-3-flash, our 4B-parameter ST-VLA exhibits superior transferability on unseen ST-Human-Planning tasks. With a 92.00\% success rate, it effectively estimates current task progress and predicts previous and next steps, indicating that that compact, domain-specific models can outperform general-purpose giants in complex temporal reasoning.

The evaluation validates that our fine-tuned ST-VLM successfully functions as a robust high-level reasoner within the hierarchical framework. By effectively distilling 3D spatial priors and 4D temporal context, the model provides a solid foundation for the subsequent low-level execution policy to master complex, long-horizon tasks in the open world.

\subsection{Evaluation of ST-VLA Capabilities}

\subsubsection{Simulation Experiments } 
\textbf{Simulation Tasks.} To evaluate \method in open-world manipulation, we conduct extensive simulation experiments on RLBench~\cite{james2020rlbench} with four representative tasks.
We use three tasks—\textit{Close Jar}, \textit{Light Bulb In}, and \textit{Put Groceries In Cupboard}—to test generalization across variations in object color, pose, and geometry. We further include the long-horizon \textit{Push Buttons} task to assess temporal reasoning and task decomposition. To assess robustness against out-of-distribution (OOD) scenarios, we partition the dataset into \textbf{Seen} and \textbf{Unseen} subsets based on variations (see Appendix~\ref{app:dataset-setup} for details). The Unseen subset contains novel colors and object shapes absent from training, posing a significant challenge to standard imitation learning. We consider two settings: \textbf{Single-task}, where a task-specific lower-level policy is trained and tested per task; and \textbf{Multi-task}, where a unified lower-level policy is trained across all four tasks to handle diverse manipulation goals.

\textbf{Baselines.} 
We report results for two \method variants: (1) \textbf{\method w/ Frozen Policy}, which uses a base policy without specialized retraining and receives VLM guidance only during inference; and (2) \textbf{\method w/ Fine-tuned (FT) Policy}, where the low-level model is trained from scratch on data processed using VLM-provided intermediate representations. All results are averaged over three random seeds. We use 3DDA~\cite{ke20243d} and 3DFA~\cite{gkanatsios20253d} as the low-level policies, and also include them as baselines to demonstrate that our proposed 3D-4D intermediate representations improve low-level policy generalization. Detailed hyperparameters are provided in Table~\ref{tab:hyperparams}.

\textbf{Analysis of Simulation Results.}
As shown in Table~\ref{tab:results}, \method consistently improves success rates for both 3DDA and 3DFA, demonstrating broad applicability. The largest gains appear in Unseen scenarios, where vanilla imitation learning models often fail under distribution shift, with our fine-tuned (FT) version improving zero-shot success rates by \textbf{44.6\%}. \method equips robots with robust semantic and spatial priors.
Even in Seen tasks, \method outperforms the baselines by simplifying the execution space. By masking task-irrelevant distractors, \method isolates the relevant spatial manifold and reduces the perceptual complexity, allowing $\pi_{lo}$ to focus exclusively on geometric execution.

Notably, the Frozen variant achieves substantial gains without specialized retraining. This validates that our smooth masks effectively focus attention while maintaining latent manifold stability. Unlike previous hierarchical methods that require joint fine-tuning, \method seamlessly integrates with existing imitation learning algorithms.

To further assess temporal reasoning, we decompose the \textit{Push Buttons} task into short-horizon primitives. Although the low-level policy $\pi_{lo}$ (trained with 3DFA) sees only single-button press episodes during training, it achieves a \textbf{93.3\% success rate} on the original long-horizon test set (sequentially pressing 2-3 buttons) when guided by \method (as shown in Table~\ref{tab:pushbutton_results}). This demonstrates that our framework effectively delegates task decomposition and sequence planning to the high-level VLM. By offloading temporal complexity, the low-level policy no longer needs to learn long-horizon trajectories, yet can execute complex multi-step sequences through modular, primitive-based control.

Finally, the full \method (FT) version exceeds the success rates of baseline models even when the latter are trained on the complete dataset. This indicates that hierarchical decomposition not only facilitates zero-shot transfer but also fundamentally raises the performance ceiling of low-level controllers in complex, multi-object environments.

\subsubsection{Real-World Experiments}

To validate real-world applicability and zero-shot generalization of \method, we conduct physical experiments using a Franka Emika Panda robot arm. We focus on objects-placing and block-stacking tasks evaluate semantic-to-spatial grounding (Figure~\ref{fig:real_env_setup}). For seen scenarios, we collect a limited set of demonstration trajectories per task to train the low-level policy.
Details of the real-world experimental setup and task definitions are provided in Appendix~\ref{app:real-world-setup}.
To rigorously assess the model's robustness, we introduce three evaluation dimensions: (1) \textbf{Zero-shot Generalization}, featuring unseen object categories and geometries; (2) \textbf{Distractor Robustness}, introducing task-irrelevant objects to clutter the workspace; and (3) \textbf{Long-horizon Chaining}, sequentially executing three consecutive manipulation stages. For the first to dimensions, we evaluate 7 distinct experimental settings with 10 trials each to provide a statistically robust performance analysis.

As illustrated in Fig.~\ref{fig:real_world_results}, \method significantly outperforms baseline 3D policies, yielding average success rate improvements of \textbf{30.3\%} in zero-shot generalization and \textbf{40.8\%} in distractor robustness. For OOD objects, ST-VLM leverages semantic knowledge to synthesize 3D guidance, compensating for the low-level policy's limited exposure. In cluttered scenes, our smooth masking suppresses task-irrelevant features, stabilizing the latent manifold and preventing the visual-noise-induced failures common in imitation learning.

In addition, \method effectively handles long-horizon tasks, achieving over  70\% success rate throughout triple-placement sequences despite being trained on short-horizon, single-action demonstrations. By offloading temporal reasoning and sub-goal planning tasks to the VLM, the framework decomposes complex multi-stage instructions into executable 3D-guided primitives. The consistent real-world performance confirms that decoupling semantic-temporal reasoning from execution via a 3D-aware interface allows robots to perform sophisticated manipulation with minimal human demonstrations.

%% file: sections/6_conclusion.tex
\section{Conclusion and Discussion}

We presented \method, a hierarchical framework bridging the perception-action gap in open-world robotic manipulation. By introducing a unified 3D-4D representation comprising 3D trajectories and cross-modally aligned smooth spatial masks, we decouple high-level semantic reasoning from low-level geometric execution. This architecture leverages latent 4D spatio-temporal context to enable both intra-action replanning and inter-action long-horizon planning. 

To support this, we introduced \methodset, a large-scale human manipulation dataset with high-fidelity 3D-4D annotations for specialized VLM fine-tuning. Experimentals in RLBench and real-world scenarios demonstrate that \method achieves superior zero-shot generalization, significantly outperforming state-of-the-art baselines on unseen tasks and cluttered environments. By offloading spatiotemporal complexity to a specialized VLM, \method provides a robust and scalable pathway toward general-purpose robot manipulation in the open world.

\textbf{Limitations and Future Work.} While \method shows robust generalization, its performance may degrade in extreme clutter where SAM2 struggles with ambiguous segmentation. Additionally, our current interface is primarily optimized for single-view execution. Future work will focus on integrating more resilient multi-view perception and adapting our hierarchical guidance to a broader range of low-level backbones, aiming to establish a truly universal and versatile framework for open-world manipulation.

%% file: main.bib
@inproceedings{zitkovich2023rt,
  title={Rt-2: Vision-language-action models transfer web knowledge to robotic control},
  author={Zitkovich, Brianna and Yu, Tianhe and Xu, Sichun and Xu, Peng and Xiao, Ted and Xia, Fei and Wu, Jialin and Wohlhart, Paul and Welker, Stefan and Wahid, Ayzaan and others},
  booktitle={Conference on Robot Learning},
  pages={2165--2183},
  year={2023},
  organization={PMLR}
}

@article{reed2022generalist,
  title={A generalist agent},
  author={Reed, Scott and Zolna, Konrad and Parisotto, Emilio and Colmenarejo, Sergio Gomez and Novikov, Alexander and Barth-Maron, Gabriel and Gimenez, Mai and Sulsky, Yury and Kay, Jackie and Springenberg, Jost Tobias and others},
  journal={arXiv preprint arXiv:2205.06175},
  year={2022}
}

@inproceedings{radford2021learning,
  title={Learning transferable visual models from natural language supervision},
  author={Radford, Alec and Kim, Jong Wook and Hallacy, Chris and Ramesh, Aditya and Goh, Gabriel and Agarwal, Sandhini and Sastry, Girish and Askell, Amanda and Mishkin, Pamela and Clark, Jack and others},
  booktitle={International conference on machine learning},
  pages={8748--8763},
  year={2021},
  organization={PmLR}
}

@article{liu2023visual,
  title={Visual instruction tuning},
  author={Liu, Haotian and Li, Chunyuan and Wu, Qingyang and Lee, Yong Jae},
  journal={Advances in neural information processing systems},
  volume={36},
  pages={34892--34916},
  year={2023}
}

@article{ahn2022can,
  title={Do as i can, not as i say: Grounding language in robotic affordances},
  author={Ahn, Michael and Brohan, Anthony and Brown, Noah and Chebotar, Yevgen and Cortes, Omar and David, Byron and Finn, Chelsea and Fu, Chuyuan and Gopalakrishnan, Keerthana and Hausman, Karol and others},
  journal={arXiv preprint arXiv:2204.01691},
  year={2022}
}

@article{liu2024moka,
  title={Moka: Open-world robotic manipulation through mark-based visual prompting},
  author={Liu, Fangchen and Fang, Kuan and Abbeel, Pieter and Levine, Sergey},
  journal={arXiv preprint arXiv:2403.03174},
  year={2024}
}

@article{ren2024grounded,
  title={Grounded sam: Assembling open-world models for diverse visual tasks},
  author={Ren, Tianhe and Liu, Shilong and Zeng, Ailing and Lin, Jing and Li, Kunchang and Cao, He and Chen, Jiayu and Huang, Xinyu and Chen, Yukang and Yan, Feng and others},
  journal={arXiv preprint arXiv:2401.14159},
  year={2024}
}

@article{jiang2022vima,
  title={Vima: General robot manipulation with multimodal prompts},
  author={Jiang, Yunfan and Gupta, Agrim and Zhang, Zichen and Wang, Guanzhi and Dou, Yongqiang and Chen, Yanjun and Fei-Fei, Li and Anandkumar, Anima and Zhu, Yuke and Fan, Linxi},
  journal={arXiv preprint arXiv:2210.03094},
  volume={2},
  number={3},
  pages={6},
  year={2022}
}

@inproceedings{bharadhwaj2024track2act,
  title={Track2act: Predicting point tracks from internet videos enables generalizable robot manipulation},
  author={Bharadhwaj, Homanga and Mottaghi, Roozbeh and Gupta, Abhinav and Tulsiani, Shubham},
  booktitle={European Conference on Computer Vision},
  pages={306--324},
  year={2024}
}

@inproceedings{shridhar2022cliport,
  title={Cliport: What and where pathways for robotic manipulation},
  author={Shridhar, Mohit and Manuelli, Lucas and Fox, Dieter},
  booktitle={Conference on robot learning},
  pages={894--906},
  year={2022},
  organization={PMLR}
}

@article{chi2025diffusion,
  title={Diffusion policy: Visuomotor policy learning via action diffusion},
  author={Chi, Cheng and Xu, Zhenjia and Feng, Siyuan and Cousineau, Eric and Du, Yilun and Burchfiel, Benjamin and Tedrake, Russ and Song, Shuran},
  journal={The International Journal of Robotics Research},
  volume={44},
  number={10-11},
  pages={1684--1704},
  year={2025},
  publisher={Sage Publications Sage UK: London, England}
}

@article{liu2022structdiffusion,
  title={Structdiffusion: Language-guided creation of physically-valid structures using unseen objects},
  author={Liu, Weiyu and Du, Yilun and Hermans, Tucker and Chernova, Sonia and Paxton, Chris},
  journal={arXiv preprint arXiv:2211.04604},
  year={2022}
}

@inproceedings{goyal2023rvt,
  title={Rvt: Robotic view transformer for 3d object manipulation},
  author={Goyal, Ankit and Xu, Jie and Guo, Yijie and Blukis, Valts and Chao, Yu-Wei and Fox, Dieter},
  booktitle={Conference on Robot Learning},
  pages={694--710},
  year={2023},
  organization={PMLR}
}

@article{driess2023palm,
  title={Palm-e: An embodied multimodal language model},
  author={Driess, Danny and Xia, Fei and Sajjadi, Mehdi SM and Lynch, Corey and Chowdhery, Aakanksha and Wahid, Ayzaan and Tompson, Jonathan and Vuong, Quan and Yu, Tianhe and Huang, Wenlong and others},
  year={2023}
}

@inproceedings{kerr2023lerf,
  title={Lerf: Language embedded radiance fields},
  author={Kerr, Justin and Kim, Chung Min and Goldberg, Ken and Kanazawa, Angjoo and Tancik, Matthew},
  booktitle={Proceedings of the IEEE/CVF international conference on computer vision},
  pages={19729--19739},
  year={2023}
}

@inproceedings{peng2023openscene,
  title={Openscene: 3d scene understanding with open vocabularies},
  author={Peng, Songyou and Genova, Kyle and Jiang, Chiyu and Tagliasacchi, Andrea and Pollefeys, Marc and Funkhouser, Thomas and others},
  booktitle={Proceedings of the IEEE/CVF conference on computer vision and pattern recognition},
  pages={815--824},
  year={2023}
}

@article{jatavallabhula2023conceptfusion,
  title={Conceptfusion: Open-set multimodal 3d mapping},
  author={Jatavallabhula, Krishna Murthy and Kuwajerwala, Alihusein and Gu, Qiao and Omama, Mohd and Chen, Tao and Maalouf, Alaa and Li, Shuang and Iyer, Ganesh and Saryazdi, Soroush and Keetha, Nikhil and others},
  journal={arXiv preprint arXiv:2302.07241},
  year={2023}
}

@article{Embodied-r1,
  title={Embodied-r1: Reinforced embodied reasoning for general robotic manipulation},
  author={Yuan, Yifu and Cui, Haiqin and Huang, Yaoting and Chen, Yibin and Ni, Fei and Dong, Zibin and Li, Pengyi and Zheng, Yan and Hao, Jianye},
  journal={arXiv preprint arXiv:2508.13998},
  year={2025}
}

@article{zhang2025peek,
  title={PEEK: Guiding and Minimal Image Representations for Zero-Shot Generalization of Robot Manipulation Policies},
  author={Zhang, Jesse and Memmel, Marius and Kim, Kevin and Fox, Dieter and Thomason, Jesse and Ramos, Fabio and B{\i}y{\i}k, Erdem and Gupta, Abhishek and Li, Anqi},
  journal={arXiv preprint arXiv:2509.18282},
  year={2025}
}

@article{ke20243d,
  title={3d diffuser actor: Policy diffusion with 3d scene representations},
  author={Ke, Tsung-Wei and Gkanatsios, Nikolaos and Fragkiadaki, Katerina},
  journal={arXiv preprint arXiv:2402.10885},
  year={2024}
}

@article{yuan2025seeing,
  title={From Seeing to Doing: Bridging Reasoning and Decision for Robotic Manipulation},
  author={Yuan, Yifu and Cui, Haiqin and Chen, Yibin and Dong, Zibin and Ni, Fei and Kou, Longxin and Liu, Jinyi and Li, Pengyi and Zheng, Yan and Hao, Jianye},
  journal={arXiv preprint arXiv:2505.08548},
  year={2025}
}

@inproceedings{huang2024manipvqa,
  title={Manipvqa: Injecting robotic affordance and physically grounded information into multi-modal large language models},
  author={Huang, Siyuan and Ponomarenko, Iaroslav and Jiang, Zhengkai and Li, Xiaoqi and Hu, Xiaobin and Gao, Peng and Li, Hongsheng and Dong, Hao},
  booktitle={2024 IEEE/RSJ International Conference on Intelligent Robots and Systems (IROS)},
  pages={7580--7587},
  year={2024},
  organization={IEEE}
}

@article{gu2023rt,
  title={Rt-trajectory: Robotic task generalization via hindsight trajectory sketches},
  author={Gu, Jiayuan and Kirmani, Sean and Wohlhart, Paul and Lu, Yao and Arenas, Montserrat Gonzalez and Rao, Kanishka and Yu, Wenhao and Fu, Chuyuan and Gopalakrishnan, Keerthana and Xu, Zhuo and others},
  journal={arXiv preprint arXiv:2311.01977},
  year={2023}
}

@article{li2025hamster,
  title={Hamster: Hierarchical action models for open-world robot manipulation},
  author={Li, Yi and Deng, Yuquan and Zhang, Jesse and Jang, Joel and Memmel, Marius and Yu, Raymond and Garrett, Caelan Reed and Ramos, Fabio and Fox, Dieter and Li, Anqi and others},
  journal={arXiv preprint arXiv:2502.05485},
  year={2025}
}

@article{zheng2024tracevla,
  title={Tracevla: Visual trace prompting enhances spatial-temporal awareness for generalist robotic policies},
  author={Zheng, Ruijie and Liang, Yongyuan and Huang, Shuaiyi and Gao, Jianfeng and Daum{\'e} III, Hal and Kolobov, Andrey and Huang, Furong and Yang, Jianwei},
  journal={arXiv preprint arXiv:2412.10345},
  year={2024}
}

@inproceedings{shi2024plug,
  title={Plug-and-play object-centric representations from “what” and “where” foundation models},
  author={Shi, Junyao and Qian, Jianing and Ma, Yecheng Jason and Jayaraman, Dinesh},
  booktitle={ICRA},
  year={2024}
}

@inproceedings{hancock2025run,
  title={Run-time observation interventions make vision-language-action models more visually robust},
  author={Hancock, Asher J and Ren, Allen Z and Majumdar, Anirudha},
  booktitle={2025 IEEE International Conference on Robotics and Automation (ICRA)},
  pages={9499--9506},
  year={2025},
  organization={IEEE}
}

@article{yuan2025roboengine,
  title={RoboEngine: Plug-and-Play Robot Data Augmentation with Semantic Robot Segmentation and Background Generation},
  author={Yuan, Chengbo and Joshi, Suraj and Zhu, Shaoting and Su, Hang and Zhao, Hang and Gao, Yang},
  journal={arXiv preprint arXiv:2503.18738},
  year={2025}
}

@article{li2025controlvla,
  title={ControlVLA: Few-shot Object-centric Adaptation for Pre-trained Vision-Language-Action Models},
  author={Li, Puhao and Wu, Yingying and Xi, Ziheng and Li, Wanlin and Huang, Yuzhe and Zhang, Zhiyuan and Chen, Yinghan and Wang, Jianan and Zhu, Song-Chun and Liu, Tengyu and others},
  journal={arXiv preprint arXiv:2506.16211},
  year={2025}
}

@article{mirjalili2025augmented,
  title={Augmented Reality for RObots (ARRO): Pointing Visuomotor Policies Towards Visual Robustness},
  author={Mirjalili, Reihaneh and J{\"u}lg, Tobias and Walter, Florian and Burgard, Wolfram},
  journal={arXiv preprint arXiv:2505.08627},
  year={2025}
}

@inproceedings{zhao2025cot,
  title={Cot-vla: Visual chain-of-thought reasoning for vision-language-action models},
  author={Zhao, Qingqing and Lu, Yao and Kim, Moo Jin and Fu, Zipeng and Zhang, Zhuoyang and Wu, Yecheng and Li, Zhaoshuo and Ma, Qianli and Han, Song and Finn, Chelsea and others},
  booktitle={Proceedings of the Computer Vision and Pattern Recognition Conference},
  pages={1702--1713},
  year={2025}
}

@article{huang2025otter,
  title={Otter: A vision-language-action model with text-aware visual feature extraction},
  author={Huang, Huang and Liu, Fangchen and Fu, Letian and Wu, Tingfan and Mukadam, Mustafa and Malik, Jitendra and Goldberg, Ken and Abbeel, Pieter},
  journal={arXiv preprint arXiv:2503.03734},
  year={2025}
}

@article{yuan2024robopoint,
  title={Robopoint: A vision-language model for spatial affordance prediction for robotics},
  author={Yuan, Wentao and Duan, Jiafei and Blukis, Valts and Pumacay, Wilbert and Krishna, Ranjay and Murali, Adithyavairavan and Mousavian, Arsalan and Fox, Dieter},
  journal={arXiv preprint arXiv:2406.10721},
  year={2024}
}

@inproceedings{parisi2022unsurprising,
  title={The unsurprising effectiveness of pre-trained vision models for control},
  author={Parisi, Simone and Rajeswaran, Aravind and Purushwalkam, Senthil and Gupta, Abhinav},
  booktitle={international conference on machine learning},
  pages={17359--17371},
  year={2022},
  organization={PMLR}
}

@article{nair2022r3m,
  title={R3m: A universal visual representation for robot manipulation},
  author={Nair, Suraj and Rajeswaran, Aravind and Kumar, Vikash and Finn, Chelsea and Gupta, Abhinav},
  journal={arXiv preprint arXiv:2203.12601},
  year={2022}
}

@article{brohan2024rt,
  title={Rt-2: Vision-language-action models transfer web knowledge to robotic control, 2023},
  author={Brohan, Anthony and Brown, Noah and Carbajal, Justice and Chebotar, Yevgen and Chen, Xi and Choromanski, Krzysztof and Ding, Tianli and Driess, Danny and Dubey, Avinava and Finn, Chelsea and others},
  journal={URL https://arxiv. org/abs/2307.15818},
  year={2024}
}

@article{kim2024openvla,
  title={Openvla: An open-source vision-language-action model},
  author={Kim, Moo Jin and Pertsch, Karl and Karamcheti, Siddharth and Xiao, Ted and Balakrishna, Ashwin and Nair, Suraj and Rafailov, Rafael and Foster, Ethan and Lam, Grace and Sanketi, Pannag and others},
  journal={arXiv preprint arXiv:2406.09246},
  year={2024}
}

@article{chen2025robohorizon,
  title={Robohorizon: An llm-assisted multi-view world model for long-horizon robotic manipulation},
  author={Chen, Zixuan and Huo, Jing and Chen, Yangtao and Gao, Yang},
  journal={arXiv preprint arXiv:2501.06605},
  year={2025}
}

@article{brohan2022rt,
  title={Rt-1: Robotics transformer for real-world control at scale},
  author={Brohan, Anthony and Brown, Noah and Carbajal, Justice and Chebotar, Yevgen and Dabis, Joseph and Finn, Chelsea and Gopalakrishnan, Keerthana and Hausman, Karol and Herzog, Alex and Hsu, Jasmine and others},
  journal={arXiv preprint arXiv:2212.06817},
  year={2022}
}

@inproceedings{chen2024spatialvlm,
  title={Spatialvlm: Endowing vision-language models with spatial reasoning capabilities},
  author={Chen, Boyuan and Xu, Zhuo and Kirmani, Sean and Ichter, Brain and Sadigh, Dorsa and Guibas, Leonidas and Xia, Fei},
  booktitle={Proceedings of the IEEE/CVF Conference on Computer Vision and Pattern Recognition},
  pages={14455--14465},
  year={2024}
}

@article{guo2025deepseek,
  title={Deepseek-r1: Incentivizing reasoning capability in llms via reinforcement learning},
  author={Guo, Daya and Yang, Dejian and Zhang, Haowei and Song, Junxiao and Zhang, Ruoyu and Xu, Runxin and Zhu, Qihao and Ma, Shirong and Wang, Peiyi and Bi, Xiao and others},
  journal={arXiv preprint arXiv:2501.12948},
  year={2025}
}

@article{yuan2025embodied,
  title={Embodied-r1: Reinforced embodied reasoning for general robotic manipulation},
  author={Yuan, Yifu and Cui, Haiqin and Huang, Yaoting and Chen, Yibin and Ni, Fei and Dong, Zibin and Li, Pengyi and Zheng, Yan and Hao, Jianye},
  journal={arXiv preprint arXiv:2508.13998},
  year={2025}
}

@inproceedings{walke2023bridgedata,
  title={Bridgedata v2: A dataset for robot learning at scale},
  author={Walke, Homer Rich and Black, Kevin and Zhao, Tony Z and Vuong, Quan and Zheng, Chongyi and Hansen-Estruch, Philippe and He, Andre Wang and Myers, Vivek and Kim, Moo Jin and Du, Max and others},
  booktitle={Conference on Robot Learning},
  pages={1723--1736},
  year={2023},
  organization={PMLR}
}

@article{khazatsky2024droid,
  title={Droid: A large-scale in-the-wild robot manipulation dataset},
  author={Khazatsky, Alexander and Pertsch, Karl and Nair, Suraj and Balakrishna, Ashwin and Dasari, Sudeep and Karamcheti, Siddharth and Nasiriany, Soroush and Srirama, Mohan Kumar and Chen, Lawrence Yunliang and Ellis, Kirsty and others},
  journal={arXiv preprint arXiv:2403.12945},
  year={2024}
}

@article{james2020rlbench,
  title={Rlbench: The robot learning benchmark \& learning environment},
  author={James, Stephen and Ma, Zicong and Arrojo, David Rovick and Davison, Andrew J},
  journal={IEEE Robotics and Automation Letters},
  volume={5},
  number={2},
  pages={3019--3026},
  year={2020},
  publisher={IEEE}
}

@article{tao2024maniskill3,
  title={Maniskill3: Gpu parallelized robotics simulation and rendering for generalizable embodied ai},
  author={Tao, Stone and Xiang, Fanbo and Shukla, Arth and Qin, Yuzhe and Hinrichsen, Xander and Yuan, Xiaodi and Bao, Chen and Lin, Xinsong and Liu, Yulin and Chan, Tse-kai and others},
  journal={arXiv preprint arXiv:2410.00425},
  year={2024}
}

@article{gkanatsios20253d,
  title={3D FlowMatch Actor: Unified 3D Policy for Single-and Dual-Arm Manipulation},
  author={Gkanatsios, Nikolaos and Xu, Jiahe and Bronars, Matthew and Mousavian, Arsalan and Ke, Tsung-Wei and Fragkiadaki, Katerina},
  journal={arXiv preprint arXiv:2508.11002},
  year={2025}
}

@inproceedings{chengravmad,
  title={GravMAD: Grounded Spatial Value Maps Guided Action Diffusion for Generalized 3D Manipulation},
  author={Chen, Yangtao and Chen, Zixuan and Yin, Junhui and Huo, Jing and Tian, Pinzhuo and Shi, Jieqi and Gao, Yang},
  booktitle={The Thirteenth International Conference on Learning Representations}
}

@inproceedings{huang2023voxposer,
  title={VoxPoser: Composable 3D Value Maps for Robotic Manipulation with Language Models},
  author={Huang, Wenlong and Wang, Chen and Zhang, Ruohan and Li, Yunzhu and Wu, Jiajun and Fei-Fei, Li},
  booktitle={Conference on Robot Learning},
  pages={540--562},
  year={2023},
  organization={PMLR}
}

@article{ray2024sat,
  title={Sat: Spatial aptitude training for multimodal language models},
  author={Ray, Arijit and Duan, Jiafei and Tan, Reuben and Bashkirova, Dina and Hendrix, Rose and Ehsani, Kiana and Kembhavi, Aniruddha and Plummer, Bryan A and Krishna, Ranjay and Zeng, Kuo-Hao and others},
  journal={arXiv e-prints},
  pages={arXiv--2412},
  year={2024}
}

@article{team2025robobrain,
  title={Robobrain 2.0 technical report},
  author={Team, BAAI RoboBrain and Cao, Mingyu and Tan, Huajie and Ji, Yuheng and Chen, Xiansheng and Lin, Minglan and Li, Zhiyu and Cao, Zhou and Wang, Pengwei and Zhou, Enshen and others},
  journal={arXiv preprint arXiv:2507.02029},
  year={2025}
}

@article{yang2025qwen3,
  title={Qwen3 technical report},
  author={Yang, An and Li, Anfeng and Yang, Baosong and Zhang, Beichen and Hui, Binyuan and Zheng, Bo and Yu, Bowen and Gao, Chang and Huang, Chengen and Lv, Chenxu and others},
  journal={arXiv preprint arXiv:2505.09388},
  year={2025}
}

@misc{openai2025gpt52,
  author       = {OpenAI},
  title        = {Introducing GPT-5.2},
  year         = {2025},
  howpublished = {\url{https://openai.com/index/introducing-gpt-5-2/}},
  note         = {Accessed: 2026-01-29},
}

@misc{google2025gemini3propreview,
  author       = {Google DeepMind},
  title        = {Gemini-3-Pro-Preview: Model Card and Preview Release},
  year         = {2025},
  howpublished = {\url{https://deepmind.google/models/gemini/pro/}},
  note         = {accessed 2026-01-29},
}

@inproceedings{lu2023vl,
  title={Vl-grasp: a 6-dof interactive grasp policy for language-oriented objects in cluttered indoor scenes},
  author={Lu, Yuhao and Fan, Yixuan and Deng, Beixing and Liu, Fangfu and Li, Yali and Wang, Shengjin},
  booktitle={2023 IEEE/RSJ International Conference on Intelligent Robots and Systems (IROS)},
  pages={976--983},
  year={2023},
  organization={IEEE}
}

@article{tong2024cambrian,
  title={Cambrian-1: A fully open, vision-centric exploration of multimodal llms},
  author={Tong, Peter and Brown, Ellis and Wu, Penghao and Woo, Sanghyun and IYER, Adithya Jairam Vedagiri and Akula, Sai Charitha and Yang, Shusheng and Yang, Jihan and Middepogu, Manoj and Wang, Ziteng and others},
  journal={Advances in Neural Information Processing Systems},
  volume={37},
  pages={87310--87356},
  year={2024}
}

@inproceedings{wang2024all,
  title={The all-seeing project v2: Towards general relation comprehension of the open world},
  author={Wang, Weiyun and Ren, Yiming and Luo, Haowen and Li, Tiantong and Yan, Chenxiang and Chen, Zhe and Wang, Wenhai and Li, Qingyun and Lu, Lewei and Zhu, Xizhou and others},
  booktitle={European Conference on Computer Vision},
  pages={471--490},
  year={2024}
}

@inproceedings{myers2015affordance,
  title={Affordance detection of tool parts from geometric features},
  author={Myers, Austin and Teo, Ching L and Ferm{\"u}ller, Cornelia and Aloimonos, Yiannis},
  booktitle={2015 IEEE international conference on robotics and automation (ICRA)},
  pages={1374--1381},
  year={2015},
  organization={IEEE}
}

@article{chen2026deco,
  title={DeCo: Task decomposition and skill composition for zero-shot generalization in long-horizon 3D manipulation},
  author={Chen, Zixuan and Yin, Junhui and Chen, Yangtao and Huo, Jing and Tian, Pinzhuo and Shi, Jieqi and Hou, Yiwen and Li, Yinchuan and Gao, Yang},
  journal={IEEE Robotics and Automation Letters},
  year={2026},
  publisher={IEEE}
}

@article{chen2024scar,
  title={Scar: Refining skill chaining for long-horizon robotic manipulation via dual regularization},
  author={Chen, Zixuan and Ji, Ze and Huo, Jing and Gao, Yang},
  journal={Advances in Neural Information Processing Systems},
  volume={37},
  pages={111679--111714},
  year={2024}
}
